\documentclass{article} 
\usepackage{collas2026_conference,times}
\usepackage{easyReview}
\usepackage{booktabs}
\usepackage{multirow}
\usepackage{algorithm}
\usepackage{algpseudocode}
\usepackage{wrapfig}
\usepackage{caption}
\usepackage[caption=false]{subfig}
\usepackage{needspace}
\usepackage{placeins}
\usepackage{listings}
\usepackage{subcaption}
\usepackage{enumitem}
\usepackage{xspace}

\lstdefinestyle{promptstyle}{
    basicstyle=\ttfamily\small,
    columns=fullflexible,
    breaklines=true,
    keepspaces=true,
    frame=single,
    showstringspaces=false,
    aboveskip=0.6\baselineskip,
    belowskip=0.6\baselineskip
}
\lstnewenvironment{promptcode}[1][]{\lstset{style=promptstyle,#1}}{}

\usepackage{xstring}
\makeatletter
\newcommand{\@autopunct}[1]{%
  \StrRight{#1}{1}[\@lastchar]%
  \IfSubStr{.,;:!?}{\@lastchar}{}{.}%
}
\renewcommand{\paragraph}[1]{\par\smallskip\textbf{#1\@autopunct{#1}}~}
\makeatother
\let\oldtodo\todo
\renewcommand{\todo}[1]{\oldtodo[inline]{#1}}
\definecolor{mdred}{rgb}{0.89, 0.10, 0.20}
\definecolor{cvprblue}{rgb}{0.21,0.49,0.74}

\newcommand{\method}{\textsc{SACK}\xspace}
\AtBeginEnvironment{equation}{\vspace{-4pt}}
\AtEndEnvironment{equation}{\vspace{-0pt}}

\usepackage{amsmath,amsfonts,bm}

\def\eqref#1{equation~\ref{#1}}

\def\1{\bm{1}}

\DeclareMathAlphabet{\mathsfit}{\encodingdefault}{\sfdefault}{m}{sl}
\SetMathAlphabet{\mathsfit}{bold}{\encodingdefault}{\sfdefault}{bx}{n}

\usepackage{hyperref}
\usepackage[nameinlink,noabbrev]{cleveref}
\hypersetup{
    colorlinks=true,
    linkcolor=red,
    filecolor=magenta,
    urlcolor=blue,
    citecolor=purple,
    pdftitle={Overleaf Example},
    pdfpagemode=FullScreen,
    }
\setkeys{Gin}{totalheight=0.82\textheight,keepaspectratio}

\title{Sequentially Acquiring Concept Knowledge to Guide Continual Learning}

\author{
Shivanand Kundargi \\ University of Maryland, Baltimore County \\ \texttt{shivank2@umbc.edu} 
\And 
Kowshik Thopalli \\ Lawrence Livermore National Laboratory \\
\texttt{thopalli1@llnl.gov}
\And
Tejas Gokhale \\ University of Maryland, Baltimore County \\
\texttt{gokhale@umbc.edu}
}

\collasfinalcopy 

\begin{document}

\maketitle

\begin{abstract}
The goal of continual learning (CL) is to adapt to new data (plasticity) while retaining the knowledge acquired from old data (stability). 
Existing methods are explicitly designed to focus on balancing stability and plasticity to mitigate the challenge of catastrophic forgetting while promoting learning. 
However, the impact of order and nature of samples from the new data that the network is trained on remains an under-explored factor. 
We argue that a CL algorithm should have the ability to rank incoming samples in terms of their relationship with prior data and their effect on the learning process. 
In this work, we investigate if scoring and prioritizing incoming data based on their semantic relationships with the model’s current knowledge can benefit CL. 
We propose \method, short for Sequentially Acquiring Concept Knowledge, a plug-and-play technique for continual learning methods. 
\method dissects categorical knowledge of the model into fine-grained concepts, computes the relationships between previously learned concepts and new concepts in each experience, and uses this relationship knowledge for prioritizing new samples. 
Experiments across several types of CL methods (regularization, replay, and prompt-based) in class-incremental and task-incremental settings demonstrate that integration of \method not only improves accuracy and reduces forgetting but also handles long-tail distribution, helps focus on semantically interpretable regions, and yields better calibrated and robust models compared to prior approaches.\\
Code: \href{https://github.com/shivanand-kundargi/SACK-CL}{https://github.com/shivanand-kundargi/SACK-CL}.
\end{abstract}

\section{Introduction}
\label{sec:intro}


The standard machine learning paradigm follows a three-step linear pipeline:
\textit{collect data $\rightarrow$ train model on data $\rightarrow$ evaluate model on data}, which is rarely a realistic scenario.
For instance, image recognition models deployed in the real world will naturally encounter input images that are different from the ones seen during training -- this \textit{distribution shift} in data could be a consequence of new objects, categories, styles, and photometric and geometric variations.
Conventional machine learning models that rely on the \iid assumption \cite{} to learn patterns in the training data often struggle to generalize to novelty in test data \citep{torralba2011unbiased,cruz2025}. 
To overcome this limitation, the field of continual learning has emerged, aiming to enable models to learn from and adapt to incoming data streams containing novel distributions.
Deep models, when sequentially trained on novel input data streams, are able to adapt to new data but at the cost of ``catastrophic'' forgetting of previous data \citep{french1999catastrophic}.

\begin{figure}[t]
    \centering
    \includegraphics[width=\linewidth]{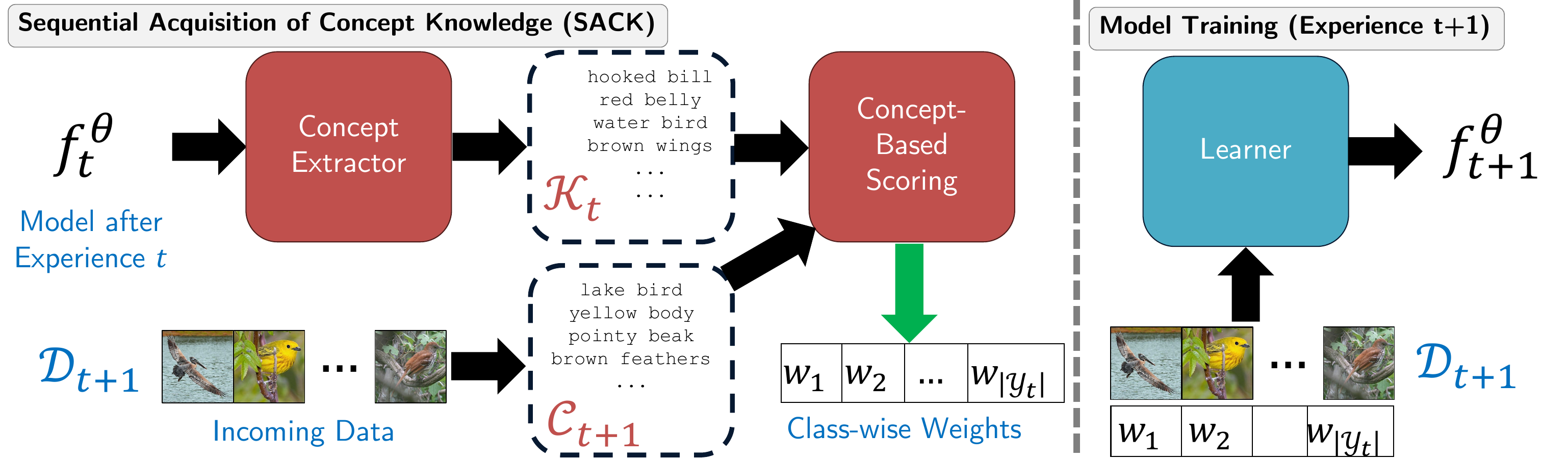}
    \caption{
    Overview of \method: Sequentially Acquiring Concept Knowledge. Given a model after experience t, we extract human-interpretable visual concepts from its internal representations and compare them with the concepts describing the incoming classes. \method computes class-wise semantic relevance scores that prioritize examples whose concepts align closely with previously learned knowledge. These concept-aligned sampling weights guide the training of experience t{+}1, forming a lightweight curriculum that improves accuracy, stability, and interpretability across continual learning streams.}
    \label{fig:teaser}
\end{figure}

Consequently, a large body of work in continual learning has focused on addressing the plasticity-stability trade-off, \ie, balancing the ability to adapt to new data (plasticity) with the retention of prior knowledge (stability) \citep{plasticity_stability1,plasticity_stability2,ER,plasticity_stability3,smith2023coda,CLSurvey1}. 
Typically, CL methods make an implicit assumption that all incoming data is uniform and has equal importance. 
However, the question of which samples are more valuable to learn from and how prioritizing them could benefit continual learning, remains largely unexplored.
This work investigates this question. 

Selective prioritization is a key part of how humans learn and how effective curricula are designed. 
\citet{vygotsky1978mind}'s 
theory of the Zone of Proximal Development  suggests that learning is most effective when new tasks lie just beyond the learner’s current competence, but are accessible with the support of existing knowledge structures.
For instance, a child who has learned to recognize circular shapes will likely find it easier to learn ovals next, while one who first learns squares may have an easier time with rectangles. 
Humans tend to build on what they already know and naturally prioritize new information that connects to familiar concepts. 
Yet, continual learning methods 
treat all incoming data equally, 
without considering how it relates to what the model existing knowledge, \ie what it had learned in prior experiences. 

The question of selective prioritization is also deeply rooted in neuroscience literature, as highlighted by \citet{stahl2015observing}, who ask: \textit{``In an environment that is dynamic and complex, how can a learner know which aspects of the world to attend to and learn from, and which to ignore?''}
Inspired by this, we argue that a CL algorithm 
should 
have the ability to rank incoming samples in terms of their relationship with prior data and their effect on the learning process. 
To that end, we introduce the \method framework (short for Sequentially Acquiring Concept Knowledge) that utilizes an interpretability technique to extract the learner's knowledge in terms of natural language concepts and then compares it with the concepts present in new data. 
This comparison guides the learner in assigning importance weights to each class in the new experience, thus allowing it to prioritize training in a way that improves both plasticity and stability. 
We perform this comparison at the concept level because concepts provide a human-aligned, interpretable abstraction of the model's knowledge, and enables the use of language models to compute semantic similarity between past and new knowledge.
By incorporating concept extraction 
and concept-based class-level importance weighting, \method departs significantly from prior CL approaches, which 
train on all incoming examples without considering their relevance to previously learned knowledge.

Our contributions and findings are summarized below:
\begin{itemize}[nosep,leftmargin=*]
    \item We introduce \method, a framework for concept-based sampling that assigns importance weights to guide training on new experiences, based on the semantic relationship between incoming classes and previously learned concepts.
    \item \method is model-, method-, and setting-agnostic, making it compatible with a wide range of CL approaches without requiring architectural changes or method-specific assumptions.
    \item We conduct extensive experiments across diverse settings including CIFAR-100, CUB-200 (fine-grained), and ImageNet-R (distribution shift) in both class-incremental (CIL) and task-incremental (TIL) settings.
    \item We demonstrate that \method, when integrated with a variety of CL approaches including regularization-, replay-, and prompt-based methods boosts plasticity, reduces forgetting, helps focus on semantically interpretable regions, leads to calibrated models, and is also effective in handling long-tailed data distributions.
\end{itemize}

\section{Related Work}
\label{sec:Related Works}
\paragraph{Continual Learning} 
The challenge of enabling models to accumulate knowledge over time without overwriting prior information \citep{french1999catastrophic} has led to extensive research in continual learning (CL), also known as lifelong or incremental learning. 
To address the stability-plasticity dilemma, three major CL strategies have emerged: regularization-based methods \citep{lwf,EWC}, replay-based techniques \citep{shin2017continual,ER}, and architectural modifications \citep{rusu2016progressive}. 
Recently, prompt-based methods \citep{smith2023coda} have gained attention, where task-specific prompts are learned while keeping the model backbone fixed. 
For further comprehensive information and detailed overview of more continual methods, we refer readers to recent literature survey studies \citep{CLSurvey1,CLsurvey2,CLsurvey3}. 

\paragraph{Curriculum Learning}
Motivated by cognitive learning theories \citep{elman1993learning}, curriculum learning explores how the ordering of training data influences learning efficiency. \citet{bengio2009curriculum} demonstrated that introducing simpler examples before harder ones can accelerate convergence and improve generalization. 
This framework closely relates to ideas from importance sampling \citep{schaul2015prioritized, zhao2015stochastic, katharopoulos2018not, baldock2021deep} and active learning \citep{AL1, ALTsurvey}, which also prioritize informative samples. 
However, curriculum learning assumes access to the full dataset in advance and is often applied in static, single-task settings.
\citet{bell2022effect} have explored the applicability of curriculum learning in continual learning by investigating the effect of task order on performance.
Our framework, \method, dynamically assesses the relative importance of each training class by probing the model’s prior semantic knowledge and uses that information to construct a suitable curriculum in a given experience. 

\paragraph{Interpretability Tools for Knowledge Discovery} 
Since our approach relies on effectively capturing the concepts learned by the model in previous experiences, research on interpretability methods is relevant to our effort. 
Classical methods like SHAP \citep{SHAP}, LIME \citep{LIME}, and Grad-CAM \citep{gradcam} offer local insights based on predictions, but fall short in capturing higher-level semantic structures embedded in the model. 
To address this shortcoming, approaches based on Sparse Autoencoders (SAEs) \citep{huben2023sparse,shu2025survey} and Sparse Transcoders \citep{paulo2025transcoders} aim to uncover the conceptual structure embedded in learned representations by enforcing sparsity and disentanglement.
Concept bottleneck models (CBMs) \citep{oikarinen2023label, koh_concept_2020} have been explored in CL to overcome catastrophic forgetting \citep{yang2024concept}, but require complex architectural changes.
Network dissection methods such as CLIP-Dissect \citep{clipdissect} utilize the vision-language embedding space of CLIP \citep{CLIP} to semantically interpret internal representations of a target network.
Given the demonstrated effectiveness of CLIP in vision tasks, we adopt CLIP-Dissect as the interpretability tool and note that \method can be integrated with other interpretability tools such as CBMs and SAEs.

\section{Sequentially Acquiring Concept Knowledge}
\label{sec:SACK}
\begin{wrapfigure}{r}{0.5\linewidth}
    \centering
    \vspace{-\intextsep}
    \includegraphics[width=\linewidth]{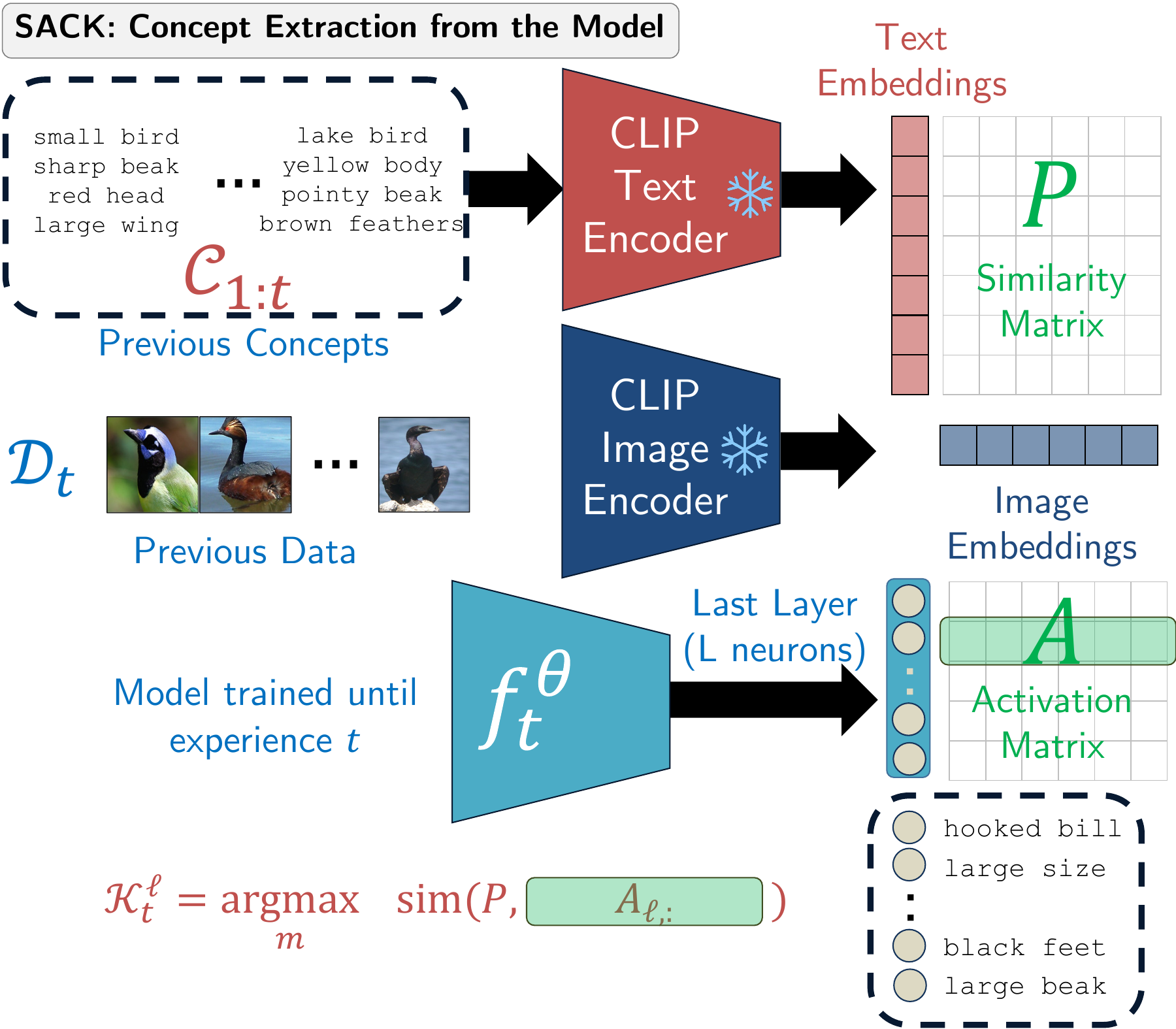}
    \caption{
    The learner has access to the accumulated natural language concept set $\mathcal{C}_{1:t}$ up to the current experience. 
    Using $\mathcal{C}_{1:t}$ and the activations from the continual learner, we obtain the activated concept set $\mathcal{K}_t^{l}$ at each experience.
    }
    \label{fig:concept_extraction_fig2}
    \vspace{-2\intextsep}
\end{wrapfigure}
The core idea of \method is to interpret concepts acquired by the model in previous experiences and those present in the current experience and then, based on this concept-level knowledge, construct a curriculum on new data to aid future learning.
We call the first step \textit{``concept extraction''} and the second step \textit{``concept-based scoring''} that we use for importance sampling.
An overview of the approach is shown in \Cref{fig:teaser}.

\subsection{Preliminaries}
\label{subsec:preliminaries}
Consider a classifier $f$ parameterized by weights $\theta$ that maps inputs $x \! \in \! \mathcal{X}$ to categorical outputs $y \! \in \! \mathcal{Y}$.
In a continual learning setting, the classifier is exposed to a sequence of \textbf{experiences} $E_t$ at discrete time steps $t \! \in \! \{1, \dots, T\}$.
Each experience $E_t$ contains data points and their categorical labels.
Model parameters $\theta$ are updated sequentially using one experience at a time.
The goal for the learner is to find the optimal $\theta$ resulting in high accuracy on the new experience $E_{t}$ while maintaining performance on all previous experiences $E_{1:t-1}$.
In this work, we consider two types of continual learning settings: class-incremental and task-incremental. 
In \textbf{Class Incremental Learning (CIL)}, 
each new experience $E_{t}$ has an associated dataset $\mathcal{D}_t$ that consists of inputs $x$ and labels $y$:
    $\mathcal{D}_t=\{x_i, y_i\}_{i=1}^{N_t}$, where $N_t$ is the number of data points for experience $D_t$.
    The set of classes contained in an experience is unique and there is no overlap with other experiences, i.e. $\mathcal{Y}_{t_1} \cap \mathcal{Y}_{t_2} = \emptyset,~\forall t_1,t_2 \! \in \! \{1, \dots, T\}$.
In \textbf{Task Incremental Learning (TIL)},
    in addition to input-label pairs, the learner can also use a task label to train the model: $\mathcal{D}_t=\{x_i, y_i, t\}_{i=1}^{N_t}$.
    The task label is available during testing, accompanying each new test sample with the information of the task in that experience, making TIL an easier setting than CIL.


\subsection{Concept Extraction: Obtaining Experience-Specific Concept Activations}
\label{subsec:Concept_Extraction}

\paragraph{Extracting Concepts from Incoming Data}
At each experience, the learner has access to the names of classes $\mathcal{Y}_t$ from that experience.
We use a large language model (e.g., GPT-4) to expand the names of classes into a set of concepts that describe each class by appropriately prompting the model as described by \citet{decider}. 
For instance, the set of concepts for the class {\it ``bear''} includes attributes such as {\it``thick coat, mighty roar, large paws, round ears, dark eyes''}, etc.
Let $\mathcal{C}_{t}$ be the set of the concepts for all classes $\mathcal{Y}_t$ in the experience $E_t$. 
Further, let $\mathcal{C}_{1:t}=\{\mathcal{C}_i\}_{i=1}^{t}$ be the combined set of concepts obtained for all experiences until $E_t$.

\paragraph{Extracting Concepts from the Model}
To extract the concepts from the model, we use a method based on CLIP-Dissect \citep{clipdissect}. As shown in \Cref{fig:concept_extraction_fig2}, we first obtain image embeddings for all $N$ images $\mathcal{X}_t$ from the experience $E_t$ using the CLIP image encoder \citep{CLIP}, and text embeddings for all $M$ concepts in $E_t$ using the CLIP text encoder.
We then compute the dot product between the image and text embeddings to obtain the matrix $P \in \mathbb{R}^{N\times M}$ where, each element of $P$, say $P_{n,m}$ represents the similarity between the $n^{th}$ image and $m^{th}$ concept.
Next, we  pass the $N$ images from experience $E_t$ to the model $f_t$ and obtain the activation matrix $A \in \mathbb{R}^{L\times N}$ where $L$ is the number of neurons in the last layer. 
Finally, we compute \textit{activated concepts,} $\mathcal{K}_{t}=\{K^{\ell}\}^{L}_{\ell=1}$ where $K^\ell$ represents the concept activated at the $\ell^{th}$ neuron of the final layer, $f_t$ which is given by:
$K^i = \arg\max_{m} \mathrm{sim}(A_{i, :}, P)$,
where $A_{i,:} \in \mathbb{R}^{1 \times N}$ represents the $i^{th}$ row of the activation matrix $A$ and $\mathrm{sim}()$ is the soft-WPMI similarity function \citep{clipdissect}.
We keep concepts with similarity scores higher than the percentile threshold $\tau$.
This filtering is performed to remove the noisy concepts and retain the ones with the highest confidence. 
For simplicity, we let $\mathcal{K}_{t}$ refer to the set of concepts after filtering.

\subsection{Concept-Based Scoring: Computing Class-Specific Weights for New Samples}
\label{subsec:concept_scoring}

\begin{wrapfigure}{r}{0.5\linewidth}
    \centering
    \vspace{-\intextsep}
    \includegraphics[width=\linewidth]{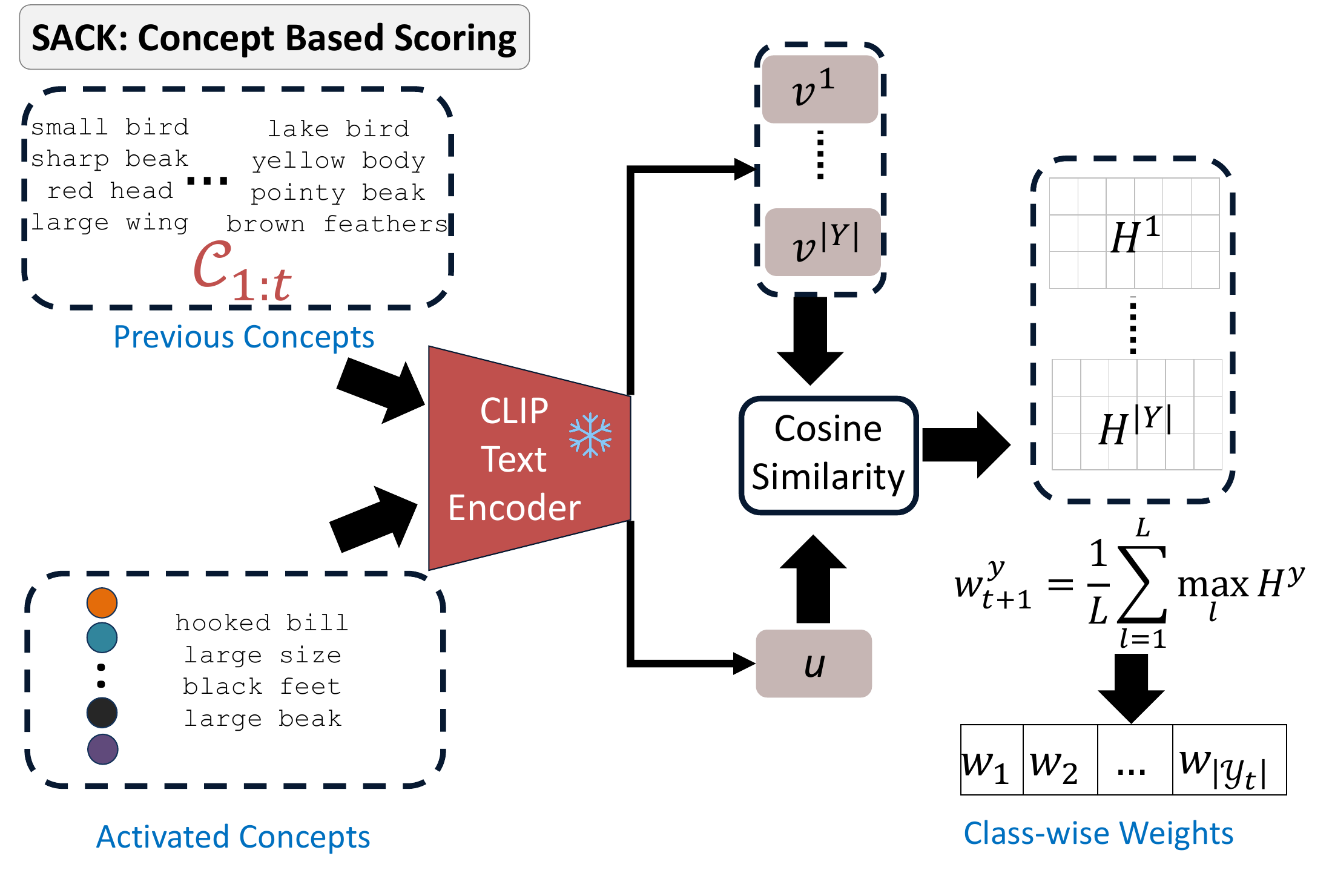}
    \caption{Per-class weights are computed by leveraging activated concepts and concepts from classes in the upcoming experience.
    }
    \label{fig:concept_scoring_fig3}
    \vspace{-2\intextsep}
\end{wrapfigure}
After obtaining concepts corresponding to the incoming experience $\mathcal{C}_{t+1} = \{ \mathcal{C}_i\}_{i=1}^{\mathcal{Y}}$ and activated concepts acquired by the model from previous experiences, $\mathcal{K}_t$, we use them to assign weights to each class $y \in \mathcal{Y}_{t+1}$ as shown in \Cref{fig:concept_scoring_fig3}.
Using the CLIP text encoder, we obtain embeddings $u \in \mathbb{R}^{768\times L}$ corresponding to $L$ activated concepts.
Similarly, we obtain embeddings $v^y \in \mathbb{R}^{768\times R}$ corresponding to $R$ concepts for each class $y \in \{1, \dots |\mathcal{Y}|\}$.
We then compute the cosine similarity between $u$ and each $v^y$ to obtain $H^y \in \mathbb{R}^{L\times R}$.
Each row $H^y_\ell$ of this similarity matrix represents the similarity of a neuron-activated concept with every concept extracted from class $y$.
We convert this similarity matrix into class-specific weights $w_{t+1}^y$ for each $y$ by aggregating the values from each row:
\begin{equation}
    \label{eq:weight-calculation}
    \begin{aligned}
        w_{t+1}^y = \frac{1}{L} \sum_{\ell=1}^{L} \max (H^y_\ell), ~\forall y \in \mathcal{Y}_{t+1}.
    \end{aligned}
\end{equation}
In \Cref{sec:aggregation}, we demonstrate that \method remains effective with combinations of multiple aggregation strategies (max, mean, and softmax), suggesting that its performance is not tied to a single aggregation choice.

\subsection{Training with Concept-Based Sampling}
\label{subsec:training}
As described in \Cref{alg:prioritized_cl}, to train on incoming experience $E_{t+1}$, we use the class-specific weights from \Cref{eq:weight-calculation} to guide the sampling of training data for each epoch within that experience.
Let $\mathbf{p}_g$ represent the class-wise sampling probabilities used by the sampler at epoch $g$, and let $\mathcal{W}$ represent the weights obtained for each class in the incoming new experience \(E_{t+1}\).
Inspired by the success and theoretical proofs of curriculum learning \citep{hacohen2019power} and importance sampling works \citep{zhao2015stochastic}, we compute sampling probabilities as:
\begin{equation}
    \label{eq:linearinterp_w2u}
        \mathbf{p}_g
        = (1-\lambda_g)\, \frac{{w}_{t+1}}{\sum_{y=1}^{\lvert \mathcal{Y}_{t+1}\rvert} w^y_{t+1}}\,\mathbf{1}
           + \lambda_g\,\frac{1}{\lvert \mathcal{Y}_{t+1}\rvert}; 
           ~~\text{where}~~
        \lambda_g = \frac{g-1}{G-1},~\forall g \in \{1,\dots,G\}.
\end{equation}
\begin{wrapfloat}{algorithm}{r}{0.55\linewidth}
    \caption{\small SACK: Sequentially Acquiring Concept Knowledge}
    \label{alg:prioritized_cl}
    \footnotesize
    \begin{algorithmic}[0]
        \Require Labeled data $D_t(\mathcal{X}_t, \mathcal{Y}_t)$ for each experience $E_t$. Model $f$ with initial parameters $\theta_0$
        
        \State \textbf{Initialize:} Experience index $t \gets 1$; parameters $\theta_1$
        \State $\theta_1 \gets \textsc{Train}(D_1, \theta_0)$
        \Comment{\textit{train model $f$ on first experience}}
        
        \For{$t \in \{2, \dots, T\}$}        
            \State \textbf{Initialize:} $\theta_t \gets \theta_{t-1}$ \Comment{\textit{update $\theta$ from previous exp}}
            
            \State $\mathcal{K}_t \gets \textsc{Dissect}(f_\theta)$ \Comment{\textit{obtain model's concepts}}
            
            \State $\mathcal{C}_t \gets \textsc{LLM}(\mathcal{Y}_t)$ \Comment{\textit{obtain new exp. class concepts}}
            
            \State $w \gets \textsc{ConceptScoring}(\mathcal{K}_t, \mathcal{C}_t)$ \Comment{\textit{obtain class-wise scores}}
            
            \For{$g = 0$ to $G$}
                \State compute $\mathbf{p}_g$ \Comment{\textit{using \Cref{eq:linearinterp_w2u}}}
                
                \State $\hat{D}_t^g \gets \textsc{WeightedSampling}(D_t, \mathbf{p}_g)$
                
                \State $\theta_t \gets \textsc{Train}(\hat{D}_t^g, \theta_t)$ \Comment{\textit{train model $f$}}
            \EndFor
        \EndFor
        \State \textbf{return} $\theta^\ast$ \Comment{\it final updated model parameters}
    \end{algorithmic}
\end{wrapfloat}

$\lambda_g$ is the interpolation coefficient that increases linearly over time, $|\mathcal{Y}_{t+1}|$ is the number of classes, and $G$ is the number of epochs.
At the beginning of each experience (epoch $g=1$), $\lambda_g=0$, \ie the sampler starts with a distribution based on weights calculated as described in \Cref{subsec:concept_scoring}.
At the end of each experience (epoch $g=G$), we have $\lambda_g=1$ and therefore the sampler ends with a distribution based on the uniform weights.

\textbf{Design Choice Rationale.}
 We design our sampling strategy to initially prioritize concept-based weights ($w$) and gradually transition toward uniform sampling. This design explicitly balances the stability-plasticity trade-off by reinforcing previously learned concepts early in training and progressively encouraging exploration of underrepresented concepts. 
Such a smooth progression defines an implicit curriculum shaped by the learner's own concept space, ensuring that at each epoch, the sampled instances are closely aligned with the current state of knowledge.

\section{Experiments}
\label{sec:Experiments}
We evaluate \method under different continual learning settings (CIL, TIL) with models of varying scales (ResNets and Vision Transformers) on \textbf{three}  benchmark datasets 
and apply it \textbf{five} different backbone methods 
representing regularization, replay, and prompt-based methods. 
Our experiments are built upon the \texttt{Mammoth} continual learning benchmark which has more than 70 methods and 20 datasets. 
\method is easily integrable with all the methods and datasets present in \texttt{Mammoth} benchmark. 
Our results show that \method not only benefits stability and plasticity but also helps to focus on semantically interpretable regions, improves long-tail classification performance, and yields better calibrated CL models.

\subsection{Experimental Setup}
In all our experiments, we use CLIP-Dissect \citep{clipdissect} to extract the concepts learned by the model from previous experiences.  
We use CLIP-Dissect in its inference phase before training starts on CL methods and we do \textbf{not} use any of its modules while training CL methods with \method. 
Hence, \method adds \textbf{negligible computational overhead} which we report for each methods and dataset\footnote{Note that we only report run-time cost, our pre-processing cost is similar to CLIP-Dissect.}.

\paragraph{Datasets}
\underline{(i) CIFAR-100} \citep{cifar100} contains $50,000$ training samples and $10,000$ test samples, each of size $32 \times 32$ belonging to 100 classes. For CL, we employ the standard practice of splitting this dataset into $10$ experiences, with each experience containing $10$ classes; \underline{(ii) CUB-200} \citep{cub200} is a fine-grained bird classification dataset comprising of $11,788$  across $200$ bird classes. This dataset serves as a challenging benchmark to demonstrate the effectiveness of \method even under the challenging fine-grained classification setting. Similar to the CIFAR-100 dataset, we split CUB-200 into 10 experiences, with each experience containing 20 classes; \underline{ (iii) ImageNet-R} \citep{imagenet_r} contains different renditions, such as artistic variations, cartoons, and sketches of images from $200$ classes of the  ImageNet \citep{imagenet_OG} dataset. ImageNet-R comprises 30,000 images ($224 \times 224$ resolution) and we split this too into $10$ experiences, with each experience containing $20$ classes.  We utilize this dataset to benchmark the robustness of \method under distribution shifts.

\begin{figure}[t]
    \centering    \includegraphics[width=\linewidth]{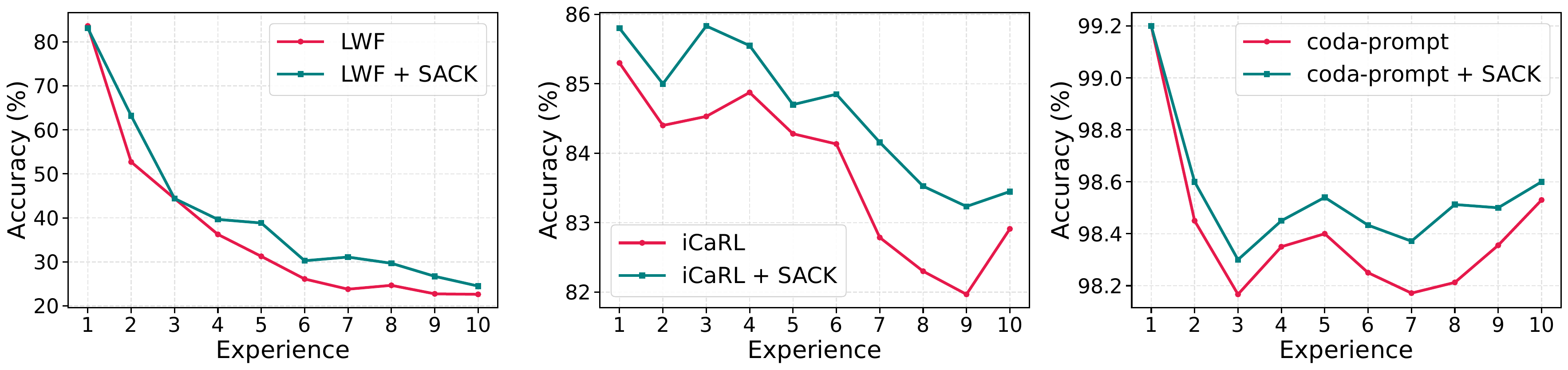}
    \caption{Task-incremental (TIL) average accuracy (Avg.ACC) across each experience for LWF, iCaRL and CODA-Prompt on CIFAR-100. \method improves average accuracy over baselines across most experiences.
    }    
    \label{fig:results_plot}
\end{figure}
\paragraph{Baseline Continual Learning Methods} Since \method is not specific to any particular choice of continual learning method, we incorporate \method in a variety of CL methods. These span regularization-based approaches such as LWF \citep{lwf}, replay-based approaches namely iCaRL \citep{rebuffi2017icarl}, DER, and DER++ \citep{der} and we also consider the recent prompting-based methods such as CODA-Prompt \citep{smith2023coda}. To integrate \method across many CL methods, we employ the {\tt mammoth} framework
which provides implementations of a host of CL methods in one repository \citep{mammoth1}. 

\paragraph{Evaluation Metrics}
To evaluate the efficacy of the proposed approach, we utilize three standard CL evaluation metrics. 
\underline{(i) Average Final Accuracy} measures the overall retention of knowledge across all experiences. It is defined as the mean accuracy computed over all experiences after training on the final experience $E_T$, ensuring that the model maintains past knowledge while learning new tasks;
\underline{(ii) Backward Transfer (BWT)}, introduced by \citet{gem} captures the effect of learning new experiences on previously learned ones. A positive BWT value indicates that learning new tasks has improved performance on earlier tasks, whereas a negative value suggests forgetting. BWT measures and quantifies the catastrophic forgetting rate.
\underline{(iii) Training Runtime(s)}: to analyze and compare the computational overhead we report average training runtime (in seconds) across all experiences on all the baseline methods and datasets.
Formally, let $A_{T,t}$ represent the accuracy of experience $E_t$ after training on $T$ experiences.
Then:
$\text{Avg.Acc} = \frac{1}{T} \sum_{t=1}^{T} A_{T,t};~ \text{BWT} = \frac{1}{T-1} \sum_{t=1}^{T-1} (A_{T,t} - A_{t,t}).$

\subsection{Implementation Details}

\paragraph{Concept Extraction.} For methods (iCaRL, DER, DER++, and LWF) that use ResNet \citep{resnet} as the backbone model, we use the last fully connected layer for dissecting and obtaining the learned experience-specific concepts. 
Similarly, for CODA-Prompt \citep{smith2023coda} which uses vision transformer backbone, we perform the network dissection on the learnable head of the vision transformer. Once the concepts along with their scores from CLIP-Dissect for each neuron are extracted, 
We prune the concept set using the $\tau=75$ percentile threshold, which leaves us with the top-$K$ concepts (with $K=25$). We also provide additional analysis by varying the percentile threshold $\tau$, showing that \method is not sensitive to the choice of $\tau$.
This is done to ensure that we only keep the concepts with the highest strength, thus effectively removing the noisy concepts. We provide a comprehensive qualitative analysis of concepts acquired at each experience and layer in the supplementary material.  

\paragraph{Class-Specific Importance Weights.} We use a weighted sampler from the PyTorch library for constructing sampler after obtaining class-specific weights as described in \Cref{subsec:concept_scoring}.
To estimate the evolving importance weights of each class for every epoch in a given experience, we follow the \Cref{eq:linearinterp_w2u} i.e, we linearly interpolate from estimated weights to uniform weights over the number of classes ($10$ for CIFAR-100 and $20$ for ImageNet-R and CUB datasets, respectively) as described in \Cref{subsec:training}. We provide a detailed analyses on different interpolation schemes and their impact in the supplementary materials. 

\paragraph{Hyperparameters.} The buffer size for the replay-based methods (iCaRL, DER, and DER++) is set to $2000$.  For all other hyperparameters and training recipes for both baseline methods with and without integration of our approach \method,  we follow the standard settings specified in {\tt mammoth library } \citep{mammoth1}. Note, we report performance of the baseline methods obtained by training them through the settings in {\tt mammoth} library to enable a fair comparison, and in most cases, we observe that  {\tt mammoth}  faithfully reproduces the results of the baseline methods reported in their respective manuscripts. Complete details of the prompts and LLM-generated concept sets for each dataset are available in our code repository and supplementary materials. All experiments were performed using NVIDIA L40S GPUs.

\begin{table*}[t]
    \centering
    \caption{Results (\%) in the class-incremental learning (CIL) setting across regularization based (LWF), replay based (iCaRL, DER, DER++) and prompt-based (CODA-Prompt) methods on CIFAR-100, CUB-200, and ImageNet-R. For each dataset, we report average accuracy (Avg.ACC), backward transfer (BWT), and runtime (s). Mean $\pm$ std is computed over 3 runs for SACK. 
    }
    \resizebox{\linewidth}{!}{
    \begin{tabular}{l ccc c ccc c ccc}
        \toprule
        \multirow{2}{*}{\textbf{Method}}
            & \multicolumn{3}{c}{\textbf{CIFAR-100}}
            & 
            & \multicolumn{3}{c}{\textbf{CUB-200}}
            & 
            & \multicolumn{3}{c}{\textbf{IMAGENET-R}} \\
        \cmidrule(lr){2-4} \cmidrule(lr){6-8} \cmidrule(lr){10-12}
            & \textbf{Avg.ACC$\uparrow$} & \textbf{BWT$\uparrow$} & \textbf{Runtime (s)$\downarrow$}
            &
            & \textbf{Avg.ACC$\uparrow$} & \textbf{BWT$\uparrow$} & \textbf{Runtime (s)$\downarrow$}
            &
            & \textbf{Avg.ACC$\uparrow$} & \textbf{BWT$\uparrow$} & \textbf{Runtime (s)$\downarrow$} \\
        \midrule

        LWF \citep{lwf}
            & 9.30 $\pm$ 0.04 & \textbf{-86.03} $\pm$ 0.14 & 64.55  && 3.58 $\pm$ 0.16 & \textbf{-24.74} $\pm$ 0.11  & 49.98 && 1.84 $\pm$ 0.37 & -23.61 $\pm$ 0.19 & 203.72 \\
        LWF \(+\) $SACK$
            & \textbf{9.32} $\pm$ 0.13 & -86.96 $\pm$ 1.78 & 65.26
            && \textbf{3.65} $\pm$ 0.24 & {-25.19} $\pm$ 0.79 & 52.16
            && \textbf{2.20} $\pm$ 0.07 & \textbf{-22.19} $\pm$ 0.78 & 205.62 \\
        \midrule

        iCaRL\citep{rebuffi2017icarl}
            & 49.35 $\pm$ 0.36 & \textbf{-17.92} $\pm$ 0.31 & 98.47 && 7.31 $\pm$ 0.56 & -7.14 $\pm$ 1.80 & 73.65 && 1.81 $\pm$ 0.20 & -49.63 $\pm$ 1.42 & 213.24 \\
        iCaRL \(+\) $SACK$
            & \textbf{50.50} $\pm$ 0.31 & -19.57 $\pm$ 0.58 & 99.27
            && \textbf{7.36} $\pm$ 0.61 & \textbf{-4.07} $\pm$ 0.72 & 75.07
            && \textbf{1.89} $\pm$ 0.11 & \textbf{-44.2} $\pm$ 0.79 & 214.14 \\
        \midrule

        DER \citep{der}
            & 48.86 $\pm$ 0.79 & -38.73 $\pm$ 2.64 & 232.95 && 6.44 $\pm$ 0.37 & \textbf{-44.61} $\pm$ 0.05 & 109.02 && 5.28 $\pm$ 0.44& -45.65 $\pm$ 1.12 & 474.59 \\
        DER \(+\) $SACK$
            & \textbf{49.89} $\pm$ 0.61 & \textbf{-36.31} $\pm$ 0.74 & 237.64
            && \textbf{6.49} $\pm$ 0.36 & {-44.94} $\pm$ 0.44 & 113.86
            && \textbf{5.79} $\pm$ 0.39 & \textbf{-44.5} $\pm$ 0.72 & 475.32 \\
        \midrule

        DER++ \citep{der}
            & 45.93 $\pm$ 0.58 & -44.86 $\pm$ 2.87 & 349.20 && 27.15 $\pm$ 0.53 & -17.26 $\pm$ 0.30 & 269.10 && 10.97 $\pm$ 0.60 & \textbf{-28.66} $\pm$ 0.96 & 859.63 \\
        DER++ \(+\) $SACK$
            & \textbf{46.78} $\pm$ 0.60 & \textbf{-43.27} $\pm$ 0.77 & 350.53
            && \textbf{27.78} $\pm$ 0.32 & \textbf{-17.14} $\pm$ 0.19 & 273.68
            && \textbf{12.53} $\pm$ 0.64 & -33.16 $\pm$ 0.82 & 860.34 \\
        \midrule

        CODA-Prompt \citep{smith2023coda}
            & 86.68 $\pm$ 0.23 & {-5.82} $\pm$ 0.07 & 195.72 && 71.35 $\pm$ 0.12 & -8.15 $\pm$ 0.43 & 53.83 && 75.24 $\pm$ 0.86 & -5.75 $\pm$ 0.61 & 397.49 \\
        CODA-Prompt \(+\) $SACK$
            & \textbf{86.69} $\pm$ 0.11 & \textbf{-5.72} $\pm$ 0.29 & 195.88
            && \textbf{72.46} $\pm$ 0.06 & \textbf{-6.98} $\pm$ 1.32 & 54.83
            && \textbf{75.65} $\pm$ 0.24 & \textbf{-5.49} $\pm$ 0.28 & 398.32 \\

        \bottomrule
    \end{tabular}
    }
    \label{tab:combined_cil}
\end{table*}
\begin{table*}[t!]
    \centering
    \huge
    \caption{Results (\%) in the task-incremental learning (TIL) setting across regularization based (LWF), replay based (iCaRL, DER \& DER++) and prompt-based (CODA-Prompt) methods  on CIFAR-100, CUB-200, and ImageNet-R. For each dataset, we report average accuracy (Avg.ACC), backward transfer (BWT), and average runtime (s) per experience. Mean $\pm$ std is computed over 3 runs for SACK 
    }
    \resizebox{\linewidth}{!}{
    \begin{tabular}{l ccc c ccc c ccc}
        \toprule
        \multirow{2}{*}{\textbf{Method}} 
            & \multicolumn{3}{c}{\textbf{CIFAR-100}} & \hphantom 
            & \multicolumn{3}{c}{\textbf{CUB-200}} & \hphantom 
            & \multicolumn{3}{c}{\textbf{IMAGENET-R}} \\
        \cmidrule(lr){2-4} \cmidrule(lr){6-8} \cmidrule(lr){10-12}
            & \textbf{Avg.ACC$\uparrow$} & \textbf{BWT$\uparrow$} & \textbf{Runtime (s)$\downarrow$} 
            && \textbf{Avg.ACC$\uparrow$} & \textbf{BWT$\uparrow$} & \textbf{Runtime (s)$\downarrow$} 
            && \textbf{Avg.ACC$\uparrow$} & \textbf{BWT$\uparrow$} & \textbf{Runtime (s)$\downarrow$} \\
        \midrule
        LWF \citep{lwf} 
            & 22.58 $\pm$ 1.73 & -71.14 $\pm$ 0.80 & 64.55 && 9.97 $\pm$ 1.64 & -20.51 $\pm$ 0.14 & 49.98 && 12.60 $\pm$ 0.19 & -11.66 $\pm$ 1.17 & 203.72 \\
        LWF \(+\) $SACK$ 
            & \textbf{24.51} $\pm$ 1.59 & \textbf{-67.23} $\pm$ 1.01 & 65.26 
            && \textbf{11.13} $\pm$ 1.24 & \textbf{-20.18} $\pm$ 0.18 & 52.16 
            && \textbf{12.83} $\pm$ 0.26 & \textbf{-11.22} $\pm$ 0.60 & 205.62 \\
        \midrule
        iCaRL \citep{rebuffi2017icarl} 
            & 82.91 $\pm$ 0.28 & -4.88 $\pm$ 0.30 & 98.47  && 26.48 $\pm$ 0.21 & 1.44 $\pm$ 1.45 & 73.65 && 15.60 $\pm$ 1.06 & \textbf{-3.22} $\pm$ 0.54 & 213.24 \\
        iCaRL \(+\) $SACK$ 
            & \textbf{83.48} $\pm$ 0.29 & \textbf{-4.16} $\pm$ 0.29 & 99.27 
            && \textbf{28.73} $\pm$ 0.22 & \textbf{4.06} $\pm$ 1.34 & 75.07 
            && \textbf{16.11} $\pm$ 0.67 & {-4.66} $\pm$ 0.50 & 214.14 \\
        \midrule
        DER \citep{der} 
            & 80.33 $\pm$ 0.55 & -10.45 $\pm$ 1.02 & 232.95 && 29.82 $\pm$ 1.18 & -18.83 $\pm$ 0.94 & 109.02 && 23.04 $\pm$ 0.72 & \textbf{-25.67} $\pm$ 2.08 & 474.59 \\
        DER \(+\) $SACK$ 
            & \textbf{82.97} $\pm$ 0.50 & \textbf{-10.07} $\pm$ 0.64 & 237.64 
            && \textbf{30.23} $\pm$ 1.04 & \textbf{-17.73} $\pm$ 0.88 & 113.86 
            && \textbf{23.55} $\pm$ 0.86 & {-26.10} $\pm$ 1.04 &  475.32\\
        \midrule
        DER++ \citep{der} 
            & 80.47 $\pm$ 0.10 & -10.97 $\pm$ 0.30 & 349.20 && 51.32 $\pm$ 0.84 & 1.03 $\pm$ 1.89 & 269.10 && 36.03 $\pm$ 1.04 & {-9.71} $\pm$ 0.77 & 859.63 \\
        DER++ \(+\) $SACK$ 
            & \textbf{81.97} $\pm$ 0.50 & \textbf{-8.58} $\pm$ 0.52 & 350.53 
            && \textbf{52.33} $\pm$ 0.78 & \textbf{2.12} $\pm$ 0.42 & 273.68 
            && \textbf{36.85} $\pm$ 0.91 & \textbf{-7.49} $\pm$ 0.54 & 860.34 \\
        \midrule
        CODA-Prompt \citep{smith2023coda} 
            & {98.56} $\pm$ 0.01 & \textbf{0.17} $\pm$ 0.02 & 195.72 && 89.01 $\pm$ 0.15 & -0.07 $\pm$ 0.04 & 53.83 && {91.21} $\pm$ 0.31 & 0.64 $\pm$ 0.06 & 397.49 \\
        CODA-Prompt \(+\) $SACK$ 
            & \textbf{98.61} $\pm$ 0.01 & 0.02 $\pm$ 0.01 &  195.88
            && \textbf{89.14} $\pm$ 0.16 & \textbf{-0.02} $\pm$ 0.09 & 54.83 
            && \textbf{91.56} $\pm$ 0.20 & \textbf{0.87} $\pm$ 0.08 & 398.32 \\
        \bottomrule
    \end{tabular}
    }
    \label{tab:combined_til}
\end{table*}
\subsection{Results}

\textbf{Class Incremental Learning}:
We evaluate the performance of \method on class-incremental learning (CIL) setting, and results are summarized in \Cref{tab:combined_cil}. We observe that  \method continues to delivers modest yet meaningful improvements across multiple baselines and datasets.
On \underline{CIFAR-100}, \method offers gains across most methods. For example, integrating \method with iCaRL improves Avg.ACC from 49.35 to 50.50 (a 1.15\% increase) and with DER++  Avg.ACC increases from 45.93 to 46.78, while registering a reduction in forgetting as BWT improves from $-44.86$ to $-43.27$. 
On \underline{CUB-200}, the benefits of \method are especially apparent. When combined with iCaRL, BWT improves from $-7.14$ to $-4.07$, demonstrating more effective retention of prior knowledge. Similar improvements are also observed in methods like DER++. 
Finally, results on the \underline{ImageNet-R} benchmark highlight \method's robustness to distribution shifts. With LWF, Avg.ACC improves from 1.84 to 2.20, and with DER++ the Avg.ACC improves by 1.56\%, and similar trends can be observed on other methods.

\textbf{Task Incremental Learning}:
\Cref{tab:combined_til} summarizes the performance of various CL methods in a task-incremental learning (TIL) setting. On \underline{CIFAR-100}, most baseline methods benefit from the integration of \method. For instance, LWF+\method relatively improves Avg.ACC by 8.55\% and shows a 5.5\% relative improvement in BWT and with \method, iCaRL achieves a 14.75\% reduction in forgetting. 
On the fine-grained \underline{CUB-200} dataset, when combined with iCaRL, Avg.ACC relatively improves by 8.5\%, and BWT turns positive, indicating strong knowledge retention. Similar improvements are observed for other methods as well. Notably, even the stronger CODA-Prompt benefits from \method, with BWT increasing from $-0.07$ to $-0.02$.
On the \underline{ImageNet-R} benchmark which introduces significant real-world distribution shifts, \method leads to similar enhancements in the performance of baseline methods. Moreover, from \Cref{fig:results_plot}, where we plot the Avg.Acc after each experience, we observe that \method outperforms baselines on most experiences. Note, these gains are realized with negligible training overhead ($<$ 1 sec) as can be seen in \Cref{tab:combined_til}.


    \textbf{Discussion:}
    From \Cref{tab:combined_cil} and \Cref{tab:combined_til}, we observe that \method's effectiveness correlates with how much of the parameters the base method actually updates. On methods that update a small set of learnable prompts parameters (like CODA-Prompt), \method's concept based prioritization has minimal parameter leverage, so gains shrink to near-zero whereas on methods that update the full backbone (replay/regularization methods like iCaRL, LWF, DER, DER++), \method is most effective and gives visible and clear gains. We inferred another interesting insight -- a closer look across the both TIL and CIL settings reveals a pattern: \method's gains are generally larger in TIL than in CIL (e.g., DER on CIFAR-100 improves by $2.64\%$ in TIL versus $1.03\%$ in CIL). We attribute this to the dominant source of error in each setting. In CIL, accuracy is limited primarily by inter-task class confusion, which sample prioritization alone cannot fully resolve, whereas in TIL the task label removes this confusion and allows the concept-driven curriculum to translate more directly into
    accuracy. We also note that gains in accuracy and forgetting are not always coupled: in a few cases (e.g., iCaRL on CIFAR-100 CIL) \method improves accuracy while incurring a small increase in forgetting, reflecting a mild reallocation of capacity toward plasticity.

\begin{figure}[t]
    \centering    
    \includegraphics[width=\linewidth]{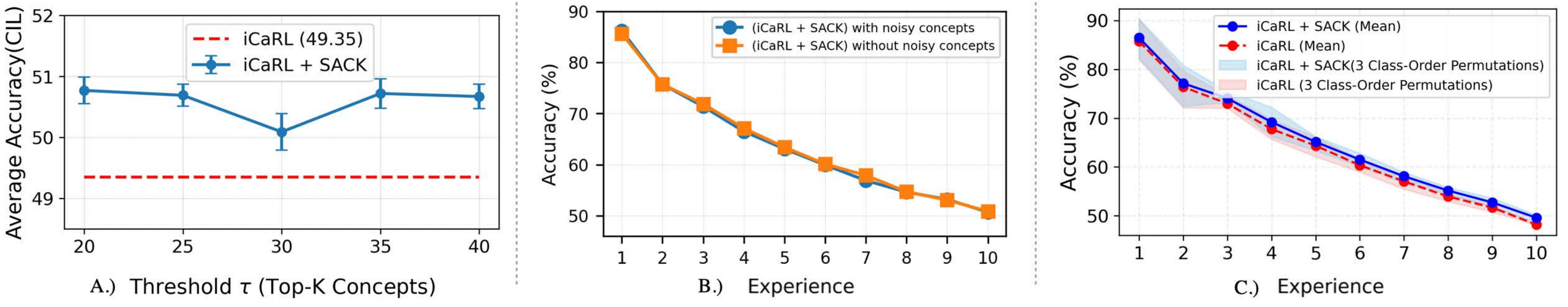}
    \caption{ Analysis of robustness of \method across diverse and challenging scenarios in the CIL setting with CIFAR-100. (A) demonstrates that \method is robust to the choice of the threshold $\tau$ which determines the number of concepts selected after performing CLIP-Dissect; (B) Performance of \method under noisy concepts, which we simulate by corrupting a subset of concepts after each experience, demonstrates that \method is resilient to concept noise. (C) Effect of perturbed class orderings across experiences. Even with shuffled class orders, \method consistently outperforms the baseline, demonstrating robustness to task ordering. We observe similar trend on other methods.
    Please refer to Appendix, for additional analysis on other models and datasets in both CIL and TIL settings.
    }    
    \label{fig:combined}
\end{figure}

\section{Analyses}
\label{sec:analyses}
In the following subsections, we present a detailed analysis of \method. We demonstrate that \method strengthens semantic grounding (\Cref{subsec:interp}), enhances robustness (\Cref{subsec:robustness}), calibration (\Cref{subsec:calibration}) in continual learners, and improves performance under long-tailed distributions (\Cref{subsec:long-tail}). We also compare \method to alternative sampling strategies (\Cref{subsec:sampling}) and examine the impact of different interpolation schemes (\Cref{subsec:interpolation_schema}).

\subsection{\method Enables CL Model to Focus on Semantically Interpretable Regions}
\label{subsec:interp}

\begin{wrapfigure}{r}{0.5\linewidth}
    \vspace{-\intextsep}
    \centering
    \includegraphics[width=\linewidth]{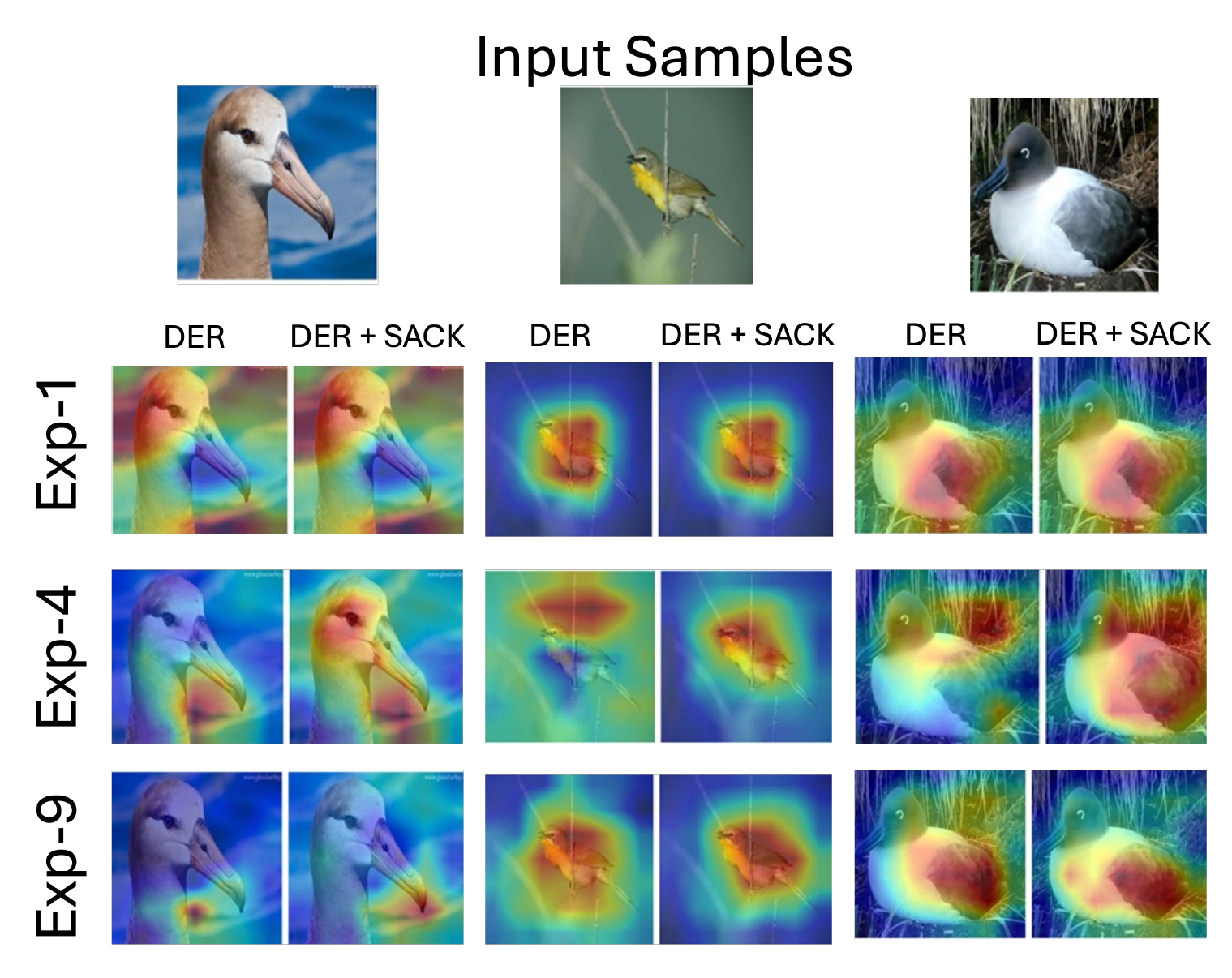}
    \caption{Semantic grounding analysis on CUB-200.}
    \vspace{-\intextsep}
    \label{fig:gradcam_viz}
\end{wrapfigure}

We perform Grad-CAM \citep{gradcam} visualization (using first experience class samples from CUB-200 dataset) on final layer model weights trained with DER \citep{der} and compare it with the visualization on final layer model weights trained with DER + \method.
This allows us to investigate if training CL models with concept guidance based framework like \method really help the models to look at semantically meaningful concepts while learning. From \Cref{fig:gradcam_viz} we observe that, for the left-most sample (black-footed-albatross), DER \citep{der} progressively drifts away from semantically interpretable regions, whereas the model trained with \method consistently focuses on semantically interpretable regions like the head and beak of the bird. We observe similar trends with the other two samples, i.e., the center (yellow-breasted-chat) and right-most (sooty-albatross). These qualitative visualizations demonstrate that \method enables the model to rely on semantically interpretable and salient regions to make its predictions.

\subsection{\method Is Robust to Threshold Choice, Concept Noise, and Class Order}
\label{subsec:robustness}

We further assess the robustness of \method to several practical sources of variation. As shown in \Cref{fig:combined} (A), \method is not sensitive to the concept-selection threshold $\tau$: across a broad range corresponding to roughly 20--40 selected concepts, iCaRL + \method consistently outperforms vanilla iCaRL while remaining largely stable. \Cref{fig:combined} (B) further shows that \method is resilient to noisy concepts; even after adding five random concepts from the \textit{Broden} concept set \citep{bau2017network_broden} to each CIFAR-100 class, performance exhibits no discernible drop, suggesting that the CLIP-Dissect thresholding step effectively filters out irrelevant concepts. Finally, \Cref{fig:combined} (C) shows that these gains do not depend on a favorable class order: across three random class orders in CIFAR-100, iCaRL + \method consistently surpasses naive iCaRL, demonstrating robustness to class-order variation as well.

\subsection{\method Leads to Better-Calibrated CL}
\label{subsec:calibration}

To assess whether \method impacts the confidence calibration of continual learners, we compute the Expected Calibration Error (ECE) \citep{ECE} on samples from the first experience, evaluated after each update with subsequent experiences.
We report the average ECE across all experiences on the CIFAR-100 test set. From \Cref{fig:calibration_ece}, we observe that \method modestly improves ECE across all logit-based continual learning baselines (DER, DER++\citep{der}, LWF\citep{lwf}, CODA-Prompt\citep{smith2023coda}). These trends highlight that \method improves accuracy while leading to either comparable or better calibration.

\subsection{\method Outperforms Different Sampling Techniques}
\label{subsec:sampling}
\begin{wraptable}{r}{0.5\linewidth}
    \vspace{-\intextsep}
    \caption{CIL performance across LWF, iCaRL, and CODA-Prompt with Uncertainty-Based (top) and Concept-Based (\method) sampling on CIFAR-100 dataset.}
    \large
    \resizebox{\linewidth}{!}{%
    \begin{tabular}{@{}lcc@{}}
        \toprule
        \textbf{Method} & \textbf{Avg. Acc (CIL) $\uparrow$} & \textbf{Avg. Acc (TIL) $\uparrow$} \\
        \midrule

        \textbf{LWF} & & \\
        \quad + Uncertainty-Based        & 9.29 $\pm$ 0.10 & 21.68 $\pm$ 0.40  \\
        \quad + Concept-Based (\method) & \textbf{9.32 $\pm$ 0.10} & \textbf{24.51 $\pm$ 1.60} \\
        \midrule

        \textbf{iCaRL} & & \\
        \quad + Uncertainty-Based        & 49.35 $\pm$ 0.30 & 82.40 $\pm$ 0.50 \\
        \quad + Concept-Based (\method) & \textbf{50.50 $\pm$ 0.30} & \textbf{83.48 $\pm$ 0.30} \\
        \midrule

        \textbf{CODA-Prompt} & & \\
        \quad + Uncertainty-Based        & 86.51 $\pm$ 0.10 & 98.44 $\pm$ 0.10 \\
        \quad + Concept-Based (\method) & \textbf{86.69 $\pm$ 0.10} & \textbf{98.56 $\pm$ 0.10} \\
        \bottomrule
    \end{tabular}%
    }
    \vspace{-\intextsep}
    \label{tab:cil_sampling_uncertainty}
\end{wraptable}

To evaluate whether \method offers advantages beyond generic sample selection heuristics, we compare \method to a standard uncertainty-based sampling strategy, where for each incoming experience, we compute class-specific prediction entropy for per sample and assign higher sampling probability to high-entropy (ie., uncertain) samples. In \Cref{tab:cil_sampling_uncertainty} we report CIL accuracy for three representative continual learning baselines (LWF, iCaRL, and CODA-Prompt) on CIFAR-100 dataset. We observe that when trained with uncertainty-based sampling versus concept-based (\method), across all three baseline methods \method yields modestly higher Avg.ACC. These gains demonstrates that concept based sampling provides more informative signal for learning as compared to heuristics based sampling methods. 
    
        
        

\subsection{\method Improves Long-tailed Classification}
\label{subsec:long-tail}

To validate the efficacy of \method in handling long-tailed distributions within CIL settings, we construct a long-tailed version of CIFAR-100 following the protocol of \citet{liu2022long}.
In this setting, the number of samples per class decays exponentially across experiences, creating a long-tail distribution.
We evaluate \method on top of three representative baselines: regularization-based LWF, replay-based iCaRL, and prompt-based CODA-Prompt with and without \method. 
From \Cref{fig:longtail-cil}, we observe that applying \method improves performance across all methods, demonstrating its ability to handle the challenges posed by long-tailed distributions in continual learning setting.


\subsubsection{Impact of Different Aggregation Strategies on \method}
\label{sec:aggregation}

\begin{wrapfigure}{r}{0.5\linewidth}
    \centering
    \vspace{-20pt}
    \includegraphics[width=\linewidth]{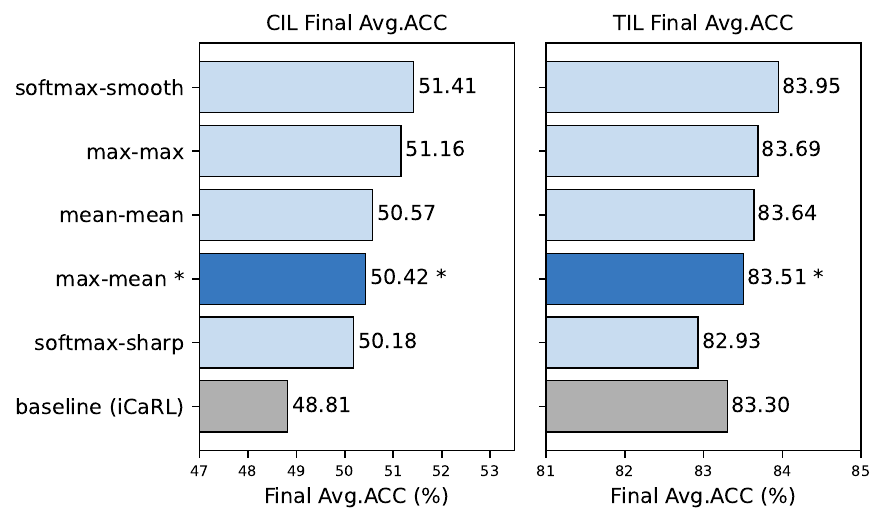}
    \caption{Effect of different aggregation strategies with iCaRL tested on the CIFAR-100 benchmark under CIL (left) and TIL (right) settings.}
    \label{fig:aggregation}
\end{wrapfigure}

From \Cref{eq:weight-calculation}, we know that $H^y \in \mathbb{R}^{L \times R}$ denotes the cosine similarity matrix for class $y$, where $L$ is the number of activated neuron-concepts, $R$ is the number of textual class-concepts, and $H^y_{l,r}$ is the similarity between activated concept $l$ and class concept $r$. We denote each aggregation rule as \textit{inner-outer}. For example, \texttt{max-mean} applies a row-wise max over the $R$ class concepts and then averages over the $L$ activated concepts to produce the class weight $w^y$. \Cref{fig:aggregation} shows that \method is robust to this design choice: most of the aggregation variants improve over the iCaRL baseline on CIL and TIL settings. In particular, softer reducers that preserve evidence from multiple high-similarity matches, most notably \texttt{softmax-smooth}, achieve the strongest final performance. 

\subsection{Effect of Different Interpolation Schedules on \method}
\label{subsec:interpolation_schema}

\begin{figure}[t]
    \centering
    \subfloat[Different interpolation schedules for \method on CIFAR-100 CIL.\label{fig:interp_schedules}]{%
        \includegraphics[width=0.32\linewidth]{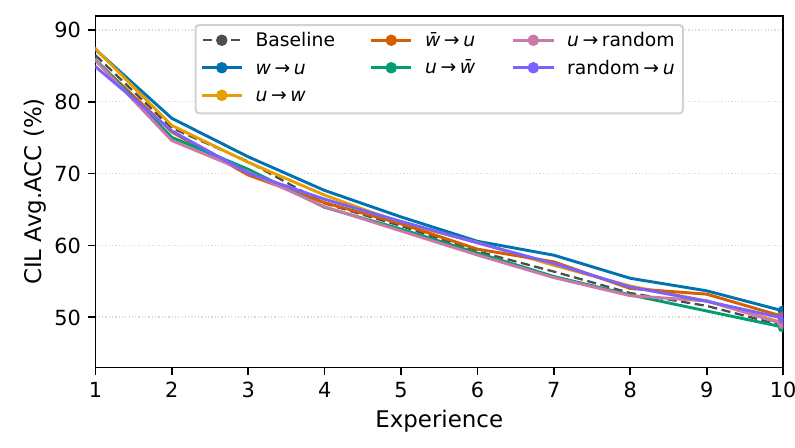}%
    }\hfill
    \subfloat[Average ECE on CIFAR-100.\label{fig:calibration_ece}]{%
        \includegraphics[width=0.32\linewidth]{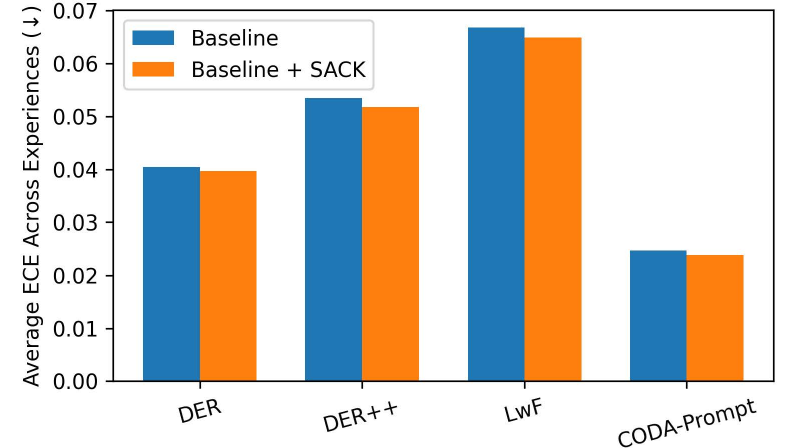}%
    }\hfill
    \subfloat[CIL performance on long-tailed CIFAR-100.\label{fig:longtail-cil}]{%
        \includegraphics[width=0.30\linewidth]{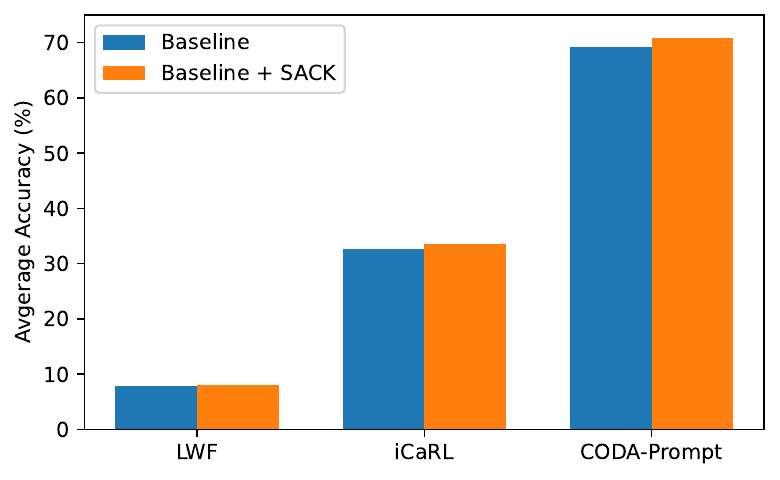}%
    }
    \caption{(a) \method performance on different interpolation schedules. (b) \method preserves or improves calibration relative to each baseline. (c) \method improves CIL performance across LWF, iCaRL, and CODA-Prompt under long-tailed class distributions.
    }
\end{figure}

In \Cref{fig:interp_schedules}, we compare several interpolation schedules for \method. The default formulation in \Cref{eq:linearinterp_w2u} is denoted by $w \rightarrow u$, where $w$ and $u$ denote the class-specific sampling weights and uniform weights, respectively. We further evaluate the alternative schedules $u \rightarrow w$, $\bar{w} \rightarrow u$, $u \rightarrow \bar{w}$, $u \rightarrow \mathrm{random}$, and $\mathrm{random} \rightarrow u$. Overall, the results suggest that $w \rightarrow u$, which emphasizes concept-similar classes early in training and gradually relaxes toward uniform sampling, yields the most favorable and comparatively better performance across experiences.
Here, $\bar{w}$ denotes the inverted weighting, $\bar{w} = 1 - w$
(renormalized over classes), which assigns the highest sampling priority to
the \emph{least} concept-similar classes, the schedule $\bar{w}\rightarrow u$
serves as a control that reverses \method's prioritization.

\section{Conclusions}
In this work, we presented a method-, model-, and setting-agnostic two-step continual learning framework that enables CL methods to prioritize samples in new experiences based on the learner's prior concept knowledge. Our results demonstrate that applying \method on top of a variety of existing methods not only enhances both plasticity and stability but also  effectively handling long-tailed distributions, yields semantically grounded and robust continual learners. \textbf{Limitations. }We acknowledge that the success of our method, \method, depends on the ability of interpretability tools to accurately assign semantic labels to neurons within the continual learner. However, as vision-language models continue to advance, we expect corresponding improvements in reliability on probed concepts. \textbf{Future Scope. }Future work includes rigorously characterizing \method theoretically from the complementary yet related perspectives and theory from continual learning, active learning and curriculum learning, as well as studying the performance of \method with interpretability tools other than CLIP-Dissect.


\section*{Acknowledgments}
This work was funded in part by the DARPA SciFy program under agreement number HR00112520301, UMBC Cybersecurity Graduate Fellows program, and SURFF award to TG.
Computing support was provided by LLNL's Livermore Computing (LC) and UMBC's HPCF.
KT's work was performed under the auspices of the U.S. Department of Energy by the Lawrence Livermore National Laboratory under Contract No. DE-AC52-07NA27344, Lawrence Livermore National Security, LLC and was supported by the LLNL-LDRD Program under Project No. 25-SI-001. 
The views, opinions, and conclusions contained herein are those of the authors and should not be interpreted as necessarily representing the official policies or endorsements, either express or implied, of employers, funding agencies, or governments.

\clearpage
\newpage
\bibliography{collas2026_conference}
\bibliographystyle{collas2026_conference}

\appendix
\clearpage

\section*{Appendix}
\section{Additional Experiments and Analyses}
\label{app:additional-experiments}
\subsection{Impact of Concept Based Regularization on SACK}
\label{app:concept-reg}

\begin{table}[h]
    \centering
    \small
    \caption{Results (\%) on CIFAR-100 in class-incremental learning (CIL) and task-incremental learning (TIL) for concept based regularization and multilayer concept probing experiments. We report average accuracy (Avg.ACC) and backward transfer (BWT). Mean $\pm$ std is computed over 3 runs for SACK where available.}
    \label{tab:concept_regularization_and_multilayer_probing}
    \begin{tabular}{@{}l cc cc@{}}
        \toprule
        \textbf{Method} &
        \multicolumn{2}{c}{\textbf{CIL}} &
        \multicolumn{2}{c}{\textbf{TIL}} \\
        \cmidrule(lr){2-3}\cmidrule(lr){4-5}
        & \textbf{Avg.ACC} $\uparrow$ & \textbf{BWT} $\uparrow$
        & \textbf{Avg.ACC} $\uparrow$ & \textbf{BWT} $\uparrow$ \\
        \midrule

        \textbf{LWF} \citep{lwf} & 9.30 & {-86.03} & 22.58 & -71.14 \\
        \quad + $SACK$ & $9.32 \pm 0.1$ & $-86.96 \pm 1.8$ & \textbf{24.51}$\pm 1.6$ & \textbf{$-67.23 \pm 1.0$} \\
        \quad + $SACK_{\text{concept\_reg}}$ & \textbf{9.44}$\pm 0.1$ & \textbf{-85.56}$\pm 0.1$ & {23.17}$\pm 1.1$ & \textbf{-64.97}$\pm 0.8$ \\
        \quad + $SACK_{\text{multiple\_layers}}$ & $9.35 \pm 0.1$ & $-86.40 \pm 1.2 $ & 23.82 $\pm 0.9$ & $-70.26 \pm 1.2$ \\
        \midrule

        \textbf{iCaRL} \citep{rebuffi2017icarl} & 49.35 & {-17.92} & 82.91 & -4.88 \\
        \quad + $SACK$ & $50.50 \pm 0.3$ & $-19.57 \pm 0.6$ & \textbf{$83.48 \pm 0.3$} & \textbf{$-4.16 \pm 0.3$} \\
        \quad + $SACK_{\text{concept\_reg}}$ & \textbf{51.23}$\pm 0.3$ & \textbf{-16.60}$\pm 0.5$ & \textbf{83.74}$\pm 0.1$ & \textbf{-4.04}$\pm 0.4$ \\
        \quad + $SACK_{\text{multiple\_layers}}$ & 50.64 $\pm 0.3$ & -19.51$\pm 0.5$ & 83.67 $\pm 0.3$ & -4.50 $\pm 0.3$ \\
        \midrule

        \textbf{DER} \citep{der} & 48.42 & -38.73 & 80.33 & -10.45 \\
        \quad + $SACK$ & $49.89 \pm 0.6$ & \textbf{$-36.31 \pm 0.7$} & \textbf{82.97}$\pm 0.5$ & \textbf{$-10.07 \pm 0.6$} \\
        \quad + $SACK_{\text{concept\_reg}}$ & \textbf{50.43}$\pm 0.6$ & \textbf{-34.49}$\pm0.6$ & 81.00$\pm0.5$ & \textbf{-9.08}$\pm0.4$ \\
        \quad + $SACK_{\text{multiple\_layers}}$ & 48.62$\pm0.5$ & -36.55$\pm0.5$ & 80.40$\pm0.6$ & -10.37$\pm0.6$ \\
        \midrule

        \textbf{DER++} \citep{der} & 45.93 & -44.86 & 80.47 & -10.97 \\
        \quad + $SACK$ & $46.78 \pm 0.6$ & \textbf{$-43.27 \pm 0.8$} & \textbf{$81.97 \pm 0.5$} & \textbf{-8.58}$\pm 0.5$ \\
        \quad + $SACK_{\text{concept\_reg}}$ & \textbf{49.40}$\pm0.6$ & \textbf{-37.47}$\pm0.5$ & 81.41$\pm0.5$ & 9.03$\pm0.3$ \\
        \quad + $SACK_{\text{multiple\_layers}}$ & 48.80$\pm0.6$ & -43.35$\pm0.8$ & 80.89$\pm0.4$ & -9.79$\pm0.3$ \\
        \midrule

        \textbf{CODA-Prompt} \citep{smith2023coda} & 86.68 & -5.82 & 98.56 & \textbf{0.17} \\
        \quad + $SACK$ & $86.69 \pm 0.1$ & \textbf{$-5.72 \pm 0.3$} & \textbf{$98.61 \pm 0.1$} & \textbf{0.02}$\pm 0.1$ \\
        \quad + $SACK_{\text{concept\_reg}}$ & \textbf{86.91}$\pm0.1$ & \textbf{-5.62}$\pm0.1$ & \textbf{98.62}$\pm0.1$ &\textbf{0.02}$\pm 0.1$  \\
        \quad + $SACK_{\text{multiple\_layers}}$ & 86.72$\pm0.1$ & -5.68$\pm0.1$ & 98.50 $\pm0.1$ & \textbf{0.02}$\pm 0.1$ \\

        \bottomrule
    \end{tabular}%
\end{table}

Towards incorporating SACK's concept based activations $\mathcal{K}_{t}$ (described in \Cref{sec:SACK}) directly into the learning objective, besides using it to compute weights for sampling -- we formulate a concept based regularization loss $L_{\text{concept\_reg}}$ as described in the equation below: 
\begin{equation}
    \label{eq:concept_reg}
    \begin{aligned}
        L_{\text{concept\_reg}} = \frac{1}{|\mathcal{K}_t|} \left( \mathcal{K}_t^{sim} \cdot \left( \frac{\mathbf{f}_{\mathcal{K}_t}(x) - \boldsymbol{\mu}}{\boldsymbol{\sigma}} \right)^2 \right)
    \end{aligned}
\end{equation}

We compute $L_{\text{concept\_reg}}$ at every step after each experience, starting from second experience $E_{2:T}$. Here, we use the thresholded concept units $\mathcal{K}_t$ (\Cref{subsec:robustness} describes the thresholding criteria), corresponding per unit soft-WPMI similarity values $\mathcal{K}_t^{sim}$, and their activations $\mathbf{f}_{\mathcal{K}_t}(x)$ in current experience $E_t$ for input $x$. The statistics $\boldsymbol{\mu}$ and $\boldsymbol{\sigma}$ represent the per unit mean and standard deviations of activations computed and cached after training on the experience $E_{t-1}$.  Hence, as shown in \Cref{eq:concept_reg}, we compute the concept based regularization $L_{\text{concept\_reg}}$ penalizing the current activation deviations from previously learned activation patterns on semantically meaningful units. We weight this standardized squared error with per unit soft-WPMI similarity scores so that units with stronger concept strength exert more influence on the gradient. We hypothesize that this formulation helps in realizing SACK's concept based sampling into a complementary regularization signal that would preserve interpretable representation throughout continual learning.
From \Cref{tab:concept_regularization_and_multilayer_probing}, where $SACK_{\text{concept\_reg}}$ indicates SACK implemented with concept based regularization on CIFAR-100 dataset, we can infer that concept based regularization benefits SACK in majority of the cases. For instance, $SACK_{\text{concept\_reg}}$ under the challenging CIL setting outperforms both naive SACK and baseline methods in terms of both average accuracy and backward transfer on all the methods. Concept based regularization also benefits TIL setting particularly on iCaRL \citep{rebuffi2017icarl} and CODA-Prompt \citep{smith2023coda} methods.

\subsection{Comparison with other concept based methods}
\label{app:cbm-methods}
\begin{wraptable}{r}{0.5\linewidth}
  \centering
  \small
  \caption{SACK performance comparison with concept-based methods on CIFAR-100 (10 experiences, CIL).}
  \label{tab:cbm_performance}
  \begin{tabular}{l c}
    \toprule
    \textbf{Method} & \textbf{Avg. Accuracy} $\uparrow$ \\
    \midrule
    IN2-LWF   & 7.34 \\
    IGCL-LWF  & 6.99 \\
    SACK-LWF  & \textbf{9.48} \\
    \bottomrule
  \end{tabular}
\end{wraptable}
We examine a prior methods that integrate semantic concepts into class-incremental settings. \citet{yang2024concept} introduce two such variants: the Intrinsically Interpretable Neural Network (IN2) and Interpretability-Guided Continual Learning (IGCL). IN2 freezes part or all of the network and applies LwF, while IGCL trains a concept bottleneck layer to map concept weights across experiences and layers various continual learning strategies on top.
These methods use concepts to mitigate catastrophic forgetting, and they are significantly different from SACK, as they do not rely on concept-based curriculum learning. We still consider these methods just to compare the continual learning performance. We compare LWF+SACK on CIFAR-100 (10 experiences) as \citet{yang2024concept} use only LWF \citep{lwf} for the IN2 variant, with 10 experiences against both variants trained with LWF \citep{lwf}, and from \Cref{tab:cbm_performance}, we observe that SACK  significantly outperforms them.

\subsection{Impact of concepts obtained by different LLMs}
\label{sup-sec:different-llms}
\begin{wraptable}{r}{0.5\linewidth}
    \vspace{-0.5\intextsep}
    \centering
    \small
    \setlength{\tabcolsep}{5pt}
    \caption{LLM comparison for iCaRL+SACK on CIFAR-100.}
    \label{tab:llm_performance}
    \begin{tabular}{l c c c}
      \toprule
      \textbf{Method} & \textbf{LLM} & \textbf{Avg. Accuracy} $\uparrow$ & \textbf{BWT} $\uparrow$ \\
      \midrule
      iCaRL+SACK & GPT-4o & \textbf{50.43} & -19.57 \\
      iCaRL+SACK & Gemini & 50.20 & \textbf{-18.58} \\
      \bottomrule
    \end{tabular}%
\end{wraptable}
To understand how the choice of large language model influences the quality of extracted concepts and consequently the performance of SACK, we investigate the impact of using different large language models (LLMs) in the concept extraction step. Specifically, we compare GPT-4o \citep{achiam2023gpt} and Gemini-Flash-2.0 \citep{comanici2025gemini} by generating concepts with each model and integrating them into SACK, evaluating the resulting systems on CIFAR-100 under the iCaRL framework. From \Cref{tab:llm_performance}, we observe that concepts derived from GPT-4o lead to consistently higher accuracy, indicating that GPT-4o provides more reliable and fine-grained semantic descriptions. Therefore, all primary experiments and analyses in the main paper use concepts extracted with GPT-4o.

\subsection{Impact of SACK with concept-extraction at different layers}
\label{app:different-layers}
\begin{wraptable}{r}{0.5\linewidth}
    \vspace{-0.5\intextsep}
    \centering
    \small
    \setlength{\tabcolsep}{5pt}
    \caption{Results (\%) showing the performance of iCaRL+SACK with concepts extracted from different intermediate layers of ResNet-50 on CIFAR-100 in the CIL setting.}
    \begin{tabular}{lccc}
        \toprule
        \textbf{Layer} & \textbf{Avg. Acc. $\uparrow$} & \textbf{BWT $\uparrow$} \\
        \midrule
        Layer 1 (conv)  & 49.53 & -19.88 \\
        Layer 2 (conv)  & 49.65 & -20.21 \\
        Last Layer (FC) & \textbf{50.43} & \textbf{-19.57} \\
        \bottomrule
    \end{tabular}%
    \label{tab:different_layers}
\end{wraptable}
We investigate the impact of extracting concepts as described in \Cref{subsec:Concept_Extraction} by considering different intermediate layers of the ResNet-50 backbone trained with iCaRL on CIFAR-100. 
We analyze concepts extracted from the convolutional layers within the residual blocks, as well as from the final fully connected layer. As shown in \Cref{tab:different_layers}, using concepts from early layers leads to a noticeable drop in performance. In contrast, concepts extracted from the final fully connected layer yield significantly better results. These findings suggest that later layers, which encode higher-level semantic information, are more suitable for concept extraction in the context of continual learning.

\subsection{Impact of SACK with concept-extraction using combination of multiple layers }
\label{app:multiple-layers}
We investigate the impact of combining extracted concept corresponding to different layers on the performance of SACK.
Towards which, we probe 4 layers of ResNet-50 backbone namely [classifier, layer1, layer2, layer3] for LWF \citep{lwf}, DER and DER++ \citep{der}, and iCaRL \citep{rebuffi2017icarl} and [last, blocks.11, blocks.10, blocks.9] from ViT \citep{ViT} projection head and backbone for CODA-Prompt. For each method, we collect thresholded concepts $\mathcal{K}_{t}$ corresponding to all 4 layers, and retain unique of those concepts combined as a total, denoted as $\mathcal{K}_{t}^{cmb}$ . Finally, we use $\mathcal{K}_{t}^{cmb}$ to compute class specific weights $w$ as shown in \Cref{eq:weight-calculation}.
From \Cref{tab:concept_regularization_and_multilayer_probing}, $SACK_{\text{multiple\_layers}}$ denotes the SACK variant that combines concepts extracted by probing multiple layers. Here, as hypothesized, we infer that combining concepts extracted by probing multiple layers benefits SACK in most of the cases. We can observe that $SACK_{\text{multiple\_layers}}$ improves the accuracy over naive SACK in most of the cases on CIFAR-100 under both TIL and CIL settings. 

\subsection{Robustness to Corruptions -- (CIFAR-100C)}
\label{sup:sec-corruptions}

\begin{figure}[h]
    \centering
    \includegraphics[width=1\linewidth]{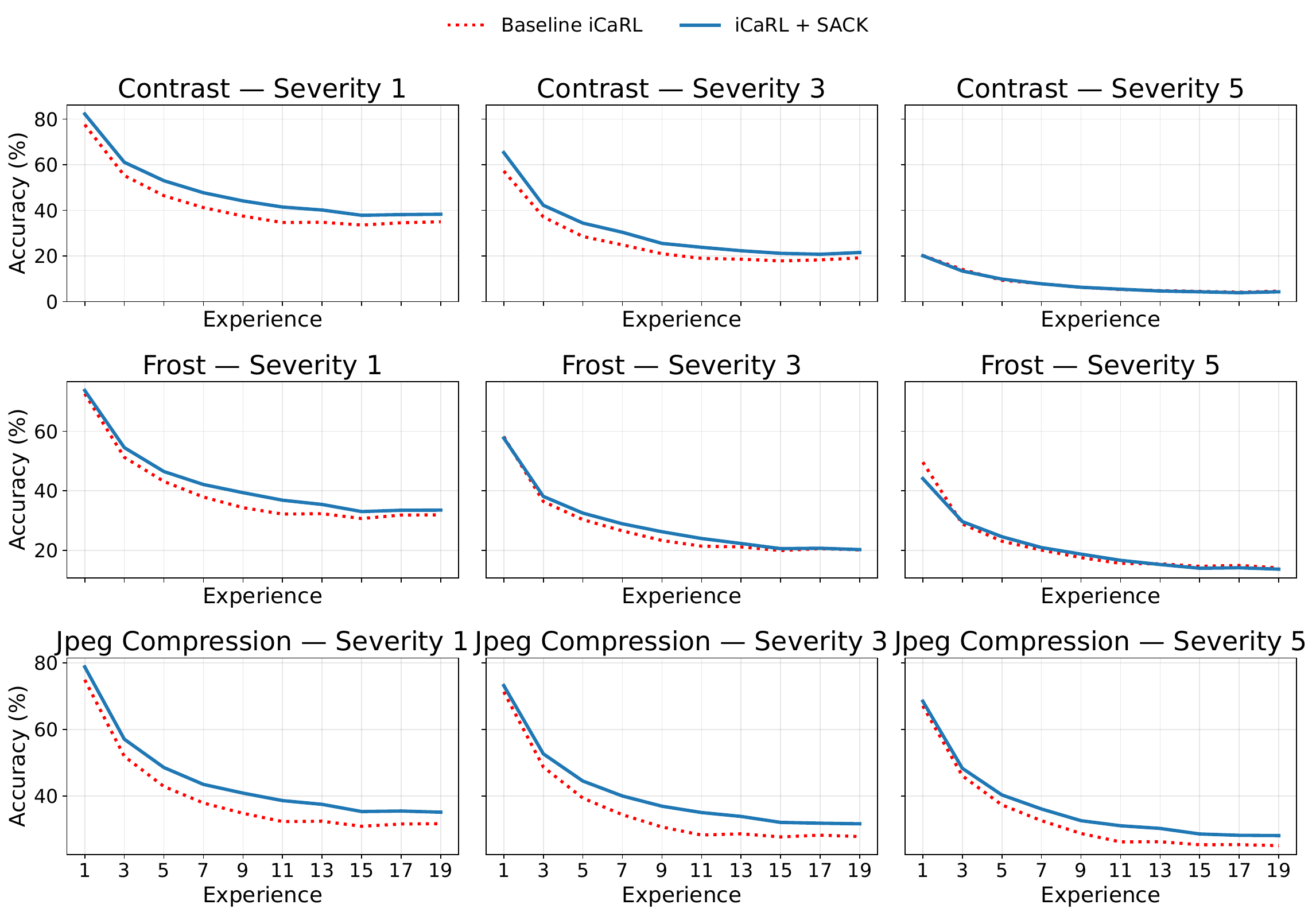}
    \caption{Comparison of iCaRL and iCaRL+SACK under \textbf{contrast, frost and jpeg compression} corruptions at severities 1 (mild), 3 (moderate), and 5 (very strong) on CIFAR-100c dataset. Across low and moderate severities, SACK maintains consistently higher accuracy and exhibits a slower degradation trend over experiences, indicating improved robustness to contrast-induced distribution shifts.
}
    \label{fig:contrast_corruption}
\end{figure}



To assess the robustness of SACK to corruptions, we evaluate one of the baseline methods(iCaRL \citep{rebuffi2017icarl}) with SACK on CIFAR-100-c \citep{hendrycks2019benchmarking} on CIL(class-incremental learning) setting. We consider contrast, frost and jpeg compression corruption types with severities 1, 3, 5. The results from \Cref{fig:contrast_corruption} show a consistent pattern across all corruptions; while both models experience the expected accuracy degradation as severity increases, SACK maintains a clear advantage over the baseline at low and moderate severity levels, often preserving 3-5 \% higher accuracy on most experiences. This indicates that concept based sampling helps retain more semantically stable representations even when inputs are perturbed. At the highest severity(level 5), where visual signal quality deteriorates sharply, both methods converge and performance of SACK remains almost similar to baseline. Across all corruptions and severities, SACK consistently performs better compared to baseline even with corruptions. This analysis highlights the effectiveness of SACK in maintaining improved performance even with corruptions.


\subsection{Thresholding (\texorpdfstring{$\tau$}{tau}) analysis on different baselines}
\label{sup:sec-threhsold-tau}
\begin{figure}[t]
    \centering
    \includegraphics[width=0.8\linewidth]{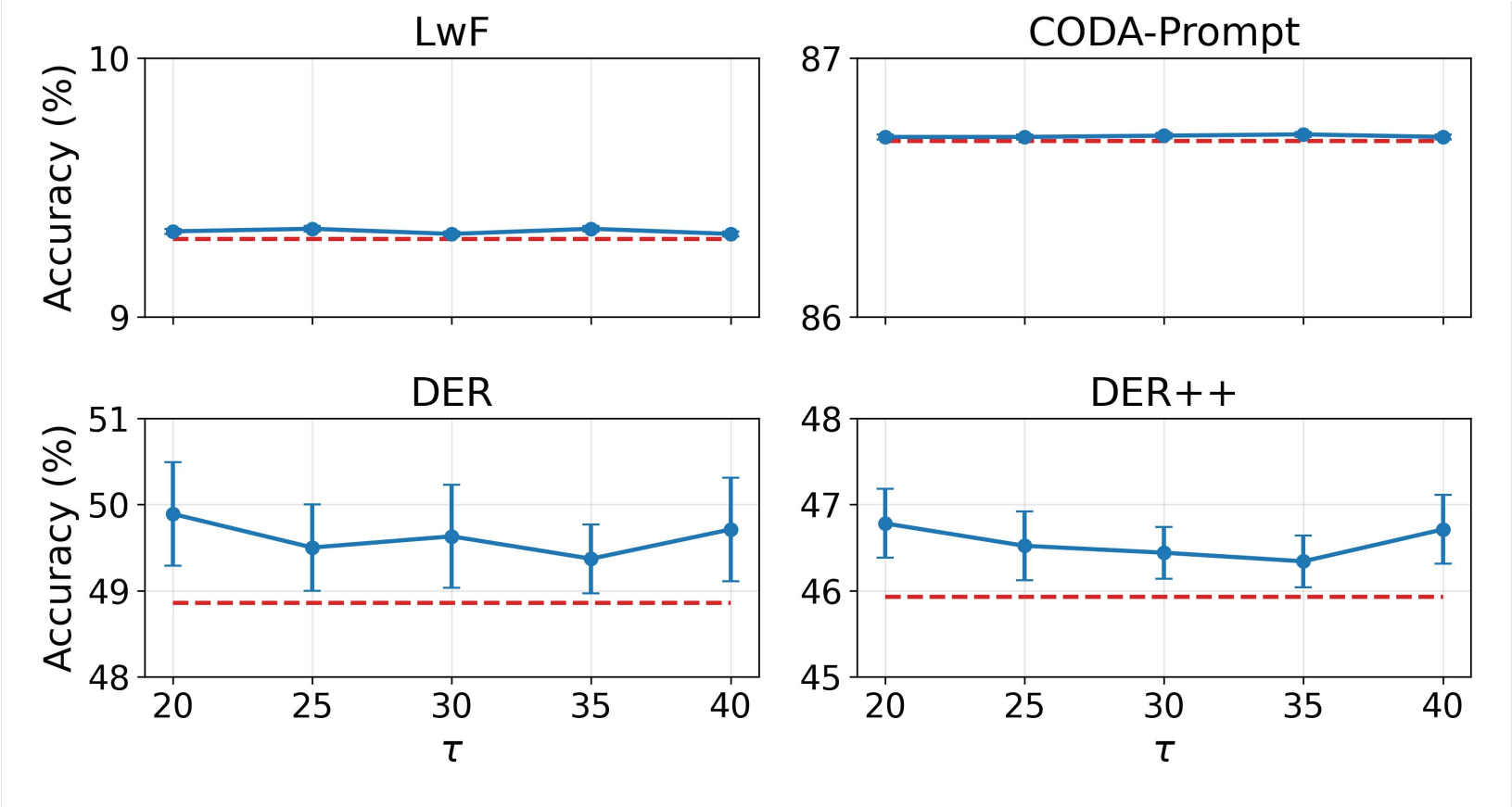}
    \caption{Effect of varying the concept-selection threshold $\tau$ on SACK with different continual learning baselines (LwF \citep{lwf}, CODA-Prompt \citep{smith2023coda}, DER and DER++ \citep{der}). For each method, we report mean CIL accuracy using SACK with $\tau$ $\in$ {20, 25, 30, 35, 40}, along with the original baseline performance shown as a dashed red line. Across all $\tau$ values, SACK remains consistently above the baseline, indicating that our method is not sensitive to the exact choice of concept threshold.}
    \label{fig:thrsholding_other_methods}
\end{figure}

In \Cref{subsec:robustness}, we demonstrated that SACK is insensitive to the choice of $\tau$ on iCaRL \citep{rebuffi2017icarl}. To further verify whether this behavior is valid on other baselines, we evaluate the effect of varying $\tau$ $\in$ {20, 25, 30, 35, 40} across four additional continual learning baselines: LWF \citep{lwf}, CODA-Prompt \citep{smith2023coda}, DER and DER++ \citep{der}. From \Cref{fig:thrsholding_other_methods}, we can observe that across all methods SACK's performance remains stable as $\tau$ changes, and overall performance of SACK is preserved regardless of concept-selection threshold. These results confirm that SACK is not sensitive to the choice of threshold across all baseline methods. 

\subsection{Order Robustness Experiments on different baselines}
\label{app:order-robustness}
\label{sup:sec-orderperm}

\begin{figure}[t]
    \centering
    \includegraphics[width=1\linewidth]{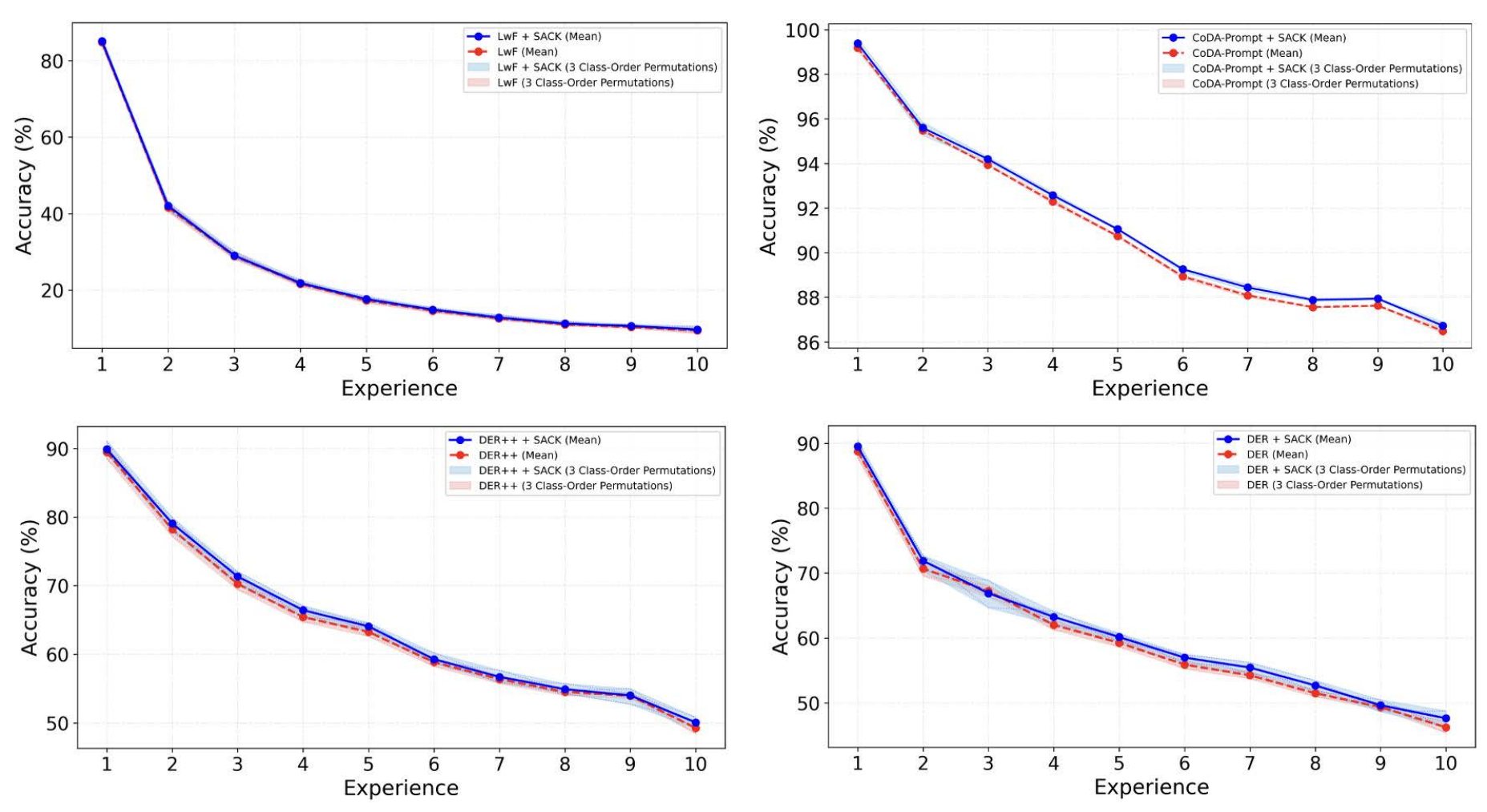}
    \caption{Order-permutation robustness across four continual learning baselines (DER \& DER++ \citep{der}, CoDA-Prompt \citep{smith2023coda}, and LwF \citep{lwf}) on CIFAR-100 under the CIL setting. Each method is evaluated with three independent class-order permutations, with shaded regions indicating variability across permutations. In all cases, SACK consistently improves both mean accuracy and stability across experiences, demonstrating that the order-robustness observed with iCaRL in the main paper generalizes across diverse CL frameworks.}
    \label{fig:permutation_order}
\end{figure}

In \Cref{subsec:robustness}, we demonstrated that SACK gives consistent gains with iCaRL \citep{rebuffi2017icarl} even when the order of classes in each experience are randomly permuted. To further examine whether this property generalizes beyond iCaRL \citep{rebuffi2017icarl}, we extend the analysis by evaluating with randomly permuted orders across a diverse set of continual learning baseline: LWF \citep{lwf}, DER \& DER++ \citep{der}, CODA-Prompt \citep{smith2023coda} on CIFAR-100 under CIL (class-incremental) setting. From \Cref{fig:permutation_order}, we observe a consistent pattern that SACK helps maintain higher mean accuracy across permutations compared to their original counterparts. This trend holds for all the baselines. These results reinforce that SACK gives consistent gains irrespective of the order of classes in each experience.

\subsection{Performance of SACK on large scale(iNaturalist) dataset}
\label{sec:inat}
\begin{wrapfigure}{r}{0.4\linewidth}
    \centering
    \vspace{-\intextsep}
    \includegraphics[width=\linewidth]{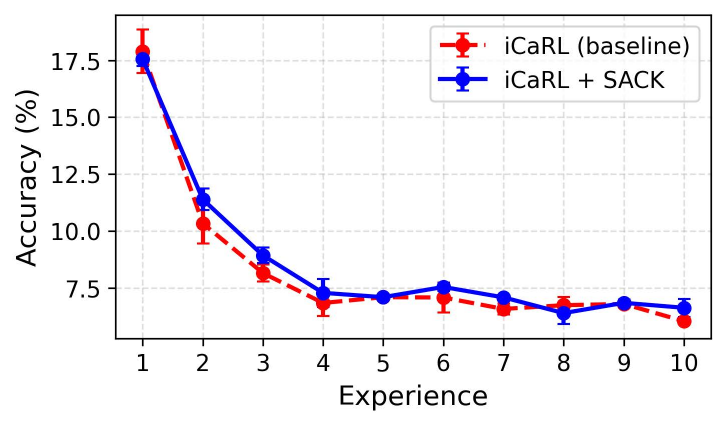}
    \caption{iNaturalist CIL results on a 300-class subset; iCaRL+SACK consistently outperforms iCaRL.}
    \label{fig:inat}
\end{wrapfigure}

To verify if the benefits of SACK persist even on large-scale, fine-grained and long-tail data distributions, we conduct an additional study on iNaturalist \citep{inat} dataset, using the first 300 classes from the full dataset.
These 300 classes contain roughly 35k highly imbalanced and fine-grained images and we divide them into 10 experiences with 30 classes each. From \Cref{fig:inat}, we can observe that SACK consistently outperforms baseline across most experiences on the challenging CIL(class-incremental) setting, indicating that its performance gains remain valid even under large-scale, fine-grained, and long-tailed data distributions.



\clearpage
\section{LLM Prompt details and Example Concepts}
\label{sec:prompt_details}
We used the following prompt to generate fine-grained visual concepts for each class. Note that, we iterate the prompt over multiple iterations and perform filtering as described in \citet{oikarinen2023label}, to obtain most relevant and plausible concepts :

\begin{promptcode}[caption={Prompt template for extracting fine-grained visual features.},label={lst:prompt-template}]
System:
"You are a helpful assistant that lists short, fine-grained visual features."

User:
"List short (1-2 word) visual fine-grained features that describe a <CATEGORY>.
Focus on its appearance, structure, texture, and environment.
Give 10-20 concepts per <CATEGORY> such that the concepts strongly define it."
\end{promptcode}

\subsection{Example Extracted Concepts}
\label{app:example-concepts}

Below we present a few examples of the visual concepts generated by the model on CIFAR-100 dataset.
\noindent\textbf{apple.}
\begin{promptcode}
"glossy surface", "smooth shiny", "yellow spots", "juicy interior",
"crunchy texture", "bright red", "round shape", ...
\end{promptcode}

\noindent\textbf{aquarium\_fish.}
\begin{promptcode}
"shimmering scales", "metallic sheen", "rainbow colors",
"striped body", "vibrant fins", "glowing eyes", ...
\end{promptcode}

\noindent\textbf{baby.}
\begin{promptcode}
"button nose", "curly hair", "pink lips", "soft skin",
"gummy smile", "rosy cheeks", ...
\end{promptcode}

\noindent\textbf{bear.}
\begin{promptcode}
"thick fur", "rounded ears", "brown coat",
"broad muzzle", "long claws", ...
\end{promptcode}

\noindent\textbf{beaver.}
\begin{promptcode}
"beady eyes", "buck teeth", "waterproof fur",
"stout body", "oval face", "rodent body", ...
\end{promptcode}

\noindent\textbf{bed.}
\begin{promptcode}
"soft pillows", "smooth sheets", "plush comforter",
"wooden slats", "inviting mattress", "cozy retreat", ...
\end{promptcode}

Full concept lists for all classes are included in the code release.


\end{document}